\documentclass[11pt]{article}

\usepackage[preprint]{acl}
\usepackage{ragged2e}
\usepackage{times}
\usepackage{latexsym}
\usepackage[T1]{fontenc}
\usepackage[utf8]{inputenc}

\usepackage{graphicx}
\usepackage{booktabs}
\usepackage{amsmath}
\usepackage{amssymb}
\usepackage{multirow}
\usepackage{xcolor}
\usepackage{subcaption}
\usepackage[accsupp]{axessibility}
\usepackage{float}
\usepackage{placeins}
\usepackage[most]{tcolorbox}

\usepackage{microtype}
\usepackage{inconsolata}

\usepackage{orcidlink}

\newtcbox{\goodmark}{
    on line,
    boxrule=0.6pt,
    colframe=green!50!black,
    colback=green!8,
    arc=1mm,
    left=0.8mm,
    right=0.8mm,
    top=0.4mm,
    bottom=0.4mm,
    boxsep=0.3mm
}
\title{A Systematic Evaluation of Positional Bias in Multi-Video Summarization with MLLMs}

\author{
  \textbf{Huangchen Xu}\textsuperscript{1},
  \textbf{Yuan Wu}\textsuperscript{1,*},
  \textbf{Yi Chang}\textsuperscript{1,2,3,*}
  \\
  \textsuperscript{1}School of Artificial Intelligence, Jilin University
  \\
  \textsuperscript{2}Engineering Research Center of Knowledge-Driven Human-Machine Intelligence, Jilin University
  \\
  \textsuperscript{3}International Center of Future Science, Jilin University
  \\
  \small{
  \href{mailto:xuhc9924@mails.jlu.edu.cn}{xuhc9924@mails.jlu.edu.cn},
  \href{mailto:yuanwu@jlu.edu.cn}{yuanwu@jlu.edu.cn},
  \href{mailto:yichang@jlu.edu.cn}{yichang@jlu.edu.cn}
  }
}

\begin{document}
\maketitle
\begin{abstract}
Multimodal Large Language Models (MLLMs) are increasingly used for video understanding, yet their reliability under multi-video inputs remains poorly understood. We study positional bias in multi-video summarization, where the quality of a per-video summary can change with the video's input slot even when the underlying content is unchanged. We construct a benchmark from ActivityNet and News videos, covering Cooking, Domestic, Leisure, and News settings with two- and four-video inputs. We evaluate nine open-source and proprietary MLLMs and measure position effects with three complementary metrics: Coverage, Directional Positional Bias (DPB), and Middle-Edge Gap (MEG). Our results show that positional effects are domain- and model-dependent: signed directional bias can be small even when middle positions underperform, and increasing visual or generation budget does not uniformly remove the imbalance. We further analyze prompt-level mitigation methods. Together, the results show that multi-video summarization remains sensitive to input protocol and position, motivating more robust order-invariant multimodal systems.

\end{abstract}

\section{Introduction}
\label{sec:intro}

Video summarization aims to condense video content into shorter yet informative representations while preserving essential information and temporal coherence~\cite{survey}. As Multimodal Large Language Models (MLLMs) become increasingly capable of processing video inputs, they have become a promising foundation for open-ended video summarization. Recent benchmarks have evaluated MLLMs on long-video understanding~\cite{zhou2025mlvubenchmarkingmultitasklong} and video summarization~\cite{jung-kim-2025-qeva}. 
Meanwhile, emerging \emph{multi-video} benchmarks have shifted attention to inputs containing multiple videos, but they primarily target understanding, perception, or reasoning rather than summarization~\cite{peng2025mvueval,bai2026mvpbenchmultivideoperceptionevaluation}. 
Less is known about \emph{multi-video summarization}, where the model must generate one aligned summary for each video in the same input.

A key concern is \textit{positional bias}, where model performance changes with input order rather than the underlying content. In NLP, large language models have been shown to favor early content in news summarization~\cite{grenander-etal-2019-countering} and to exhibit ``lost in the middle'' behavior in long-context settings~\cite{lost_in_the_middle}. Recent multimodal work has also examined positional effects in multi-image understanding~\cite{tian2025identifyingmitigatingpositionbias} and video probing benchmarks~\cite{xia2025videolevelgaugeinvestigatingcontextualpositional}. Yet these protocols mainly test whether models can retrieve or use target evidence after its position changes. Multi-video summarization poses a different challenge: the model must bind each summary to the correct video while balancing limited summary detail across multiple input slots.

In this work, we present a systematic study of positional bias in multi-video summarization. We construct a benchmark from ActivityNet~\cite{caba2015activitynet} and the News Video Dataset~\cite{whitehead-etal-2018-incorporating}, covering four scenario categories: \textit{Cooking}, \textit{Domestic}, \textit{Leisure}, and \textit{News}. We evaluate both pairwise and four-video inputs using cyclic orderings, so that each video appears at each position exactly once. Because lexical-overlap metrics are insufficient for this highly abstractive task, we use dataset-adaptive and reference-based Coverage: ActivityNet-derived domains are scored with an LLM-as-a-Judge protocol, while News uses extractive reference-fragment coverage. We then report Coverage, Directional Positional Bias (DPB), and Middle-Edge Gap (MEG) to distinguish overall information preservation, earlier-versus-later preference, and middle-position weakness. Anonymous code and data are available at \url{https://anonymous.4open.science/r/annoym07}.

Our contributions are as follows:
\begin{enumerate}
    \item \textbf{Benchmark and evaluation protocol}: We construct a benchmark over \textit{Cooking}, \textit{Domestic}, \textit{Leisure}, and \textit{News}. We also introduce an evaluation protocol that measures Coverage, Directional Positional Bias (DPB), and Middle-Edge Gap (MEG), separating earlier-versus-later preference from middle-position weakness.

    \item \textbf{Model- and domain-dependent positional effects}: We evaluate nine open-source and proprietary MLLMs and show that positional effects are model- and domain-dependent. Several settings have near-zero DPB but negative MEG, which would be missed by directional bias alone.

    \item \textbf{Robustness checks and mitigation analysis}: We examine visual budget, requested summary length, boundary format, prompt placement, and mitigation to analyze position effects.
\end{enumerate}

\section{Related Work}
\label{sec:related}

\subsection{Positional Bias}

Positional bias refers to models' sensitivity to where information or candidates appear in the input, beyond their underlying content. In text settings, prior work has studied lead or position bias in summarization~\cite{grenander-etal-2019-countering,schilcher-etal-2025-characterizing}, the ``lost in the middle'' effect in long-context QA and retrieval~\cite{lost_in_the_middle}, and position-dependent faithfulness in long-form summarization~\cite{wan-etal-2025-positional}. Similar order effects also appear when LLMs are used as judges, rankers, or recommenders~\cite{wang-etal-2024-large-language-models-fair,shi2025judgingjudgessystematicstudy,koo-etal-2024-benchmarking,hou2024largelanguagemodelszeroshot}. These studies motivate a range of mitigation strategies, from training or attention-based methods to lightweight inference-time controls such as order perturbation and attention-guiding prompts~\cite{wang-etal-2024-large-language-models-fair,NEURIPS2024_6ffdbbe3,wan-etal-2025-positional,tian2025identifyingmitigatingpositionbias}.

In multimodal settings, recent work has begun to identify analogous effects. \citet{tian2025identifyingmitigatingpositionbias} show that reordering images can substantially affect multi-image reasoning, while Video-LevelGauge~\cite{xia2025videolevelgaugeinvestigatingcontextualpositional} studies video LLMs through probing tasks that place relevant visual evidence at different positions. Our work differs in both task structure and evaluation target. Rather than testing whether a model retrieves localized evidence from a moved probe, we study whether it maintains balanced information coverage when generating aligned summaries for multiple videos under different input orders.

\subsection{Automatic Evaluation for Summarization}

Automatic summarization evaluation has long relied on lexical-overlap metrics
such as BLEU, ROUGE, and METEOR
\cite{evaluation1,papineni-etal-2002-bleu,lin-2004-rouge,banerjee-lavie-2005-meteor},
and later embedding-based metrics such as BERTScore~\cite{bert-score}. These metrics are inexpensive and scalable, but they mainly measure text similarity and do not directly assess whether the generated summary preserves or aligns with specific information in the reference.

Recent work has explored more semantic evaluation protocols. LLM-based
evaluators such as GPTScore and G-Eval use large language models to judge
generated text along flexible task-specific criteria
\cite{fu-etal-2024-gptscore,liu-etal-2023-g}. QA-based metrics such as QuestEval
evaluate summarization through question generation and question answering
\cite{rebuffel-etal-2021-data}, and recent video-summary evaluation further
uses multimodal QA to assess coverage, factuality, and chronology
\cite{jung-kim-2025-qeva}. These approaches provide richer diagnostics than
surface overlap, but they can require multiple model calls per summary and are
costly for large-scale position-wise evaluation over many videos, orders, and
slots.

For short news summaries, extractive-fragment analysis provides another useful
perspective by measuring partial phrase-level overlap~\cite{grusky-etal-2018-newsroom}. This
line of work is especially relevant when references are concise and contain
fixed named entities or event phrases.
\section{Evaluation}
\label{sec:evaluation}

\begin{figure*}[t]
  \centering
  \includegraphics[width=0.95\textwidth]{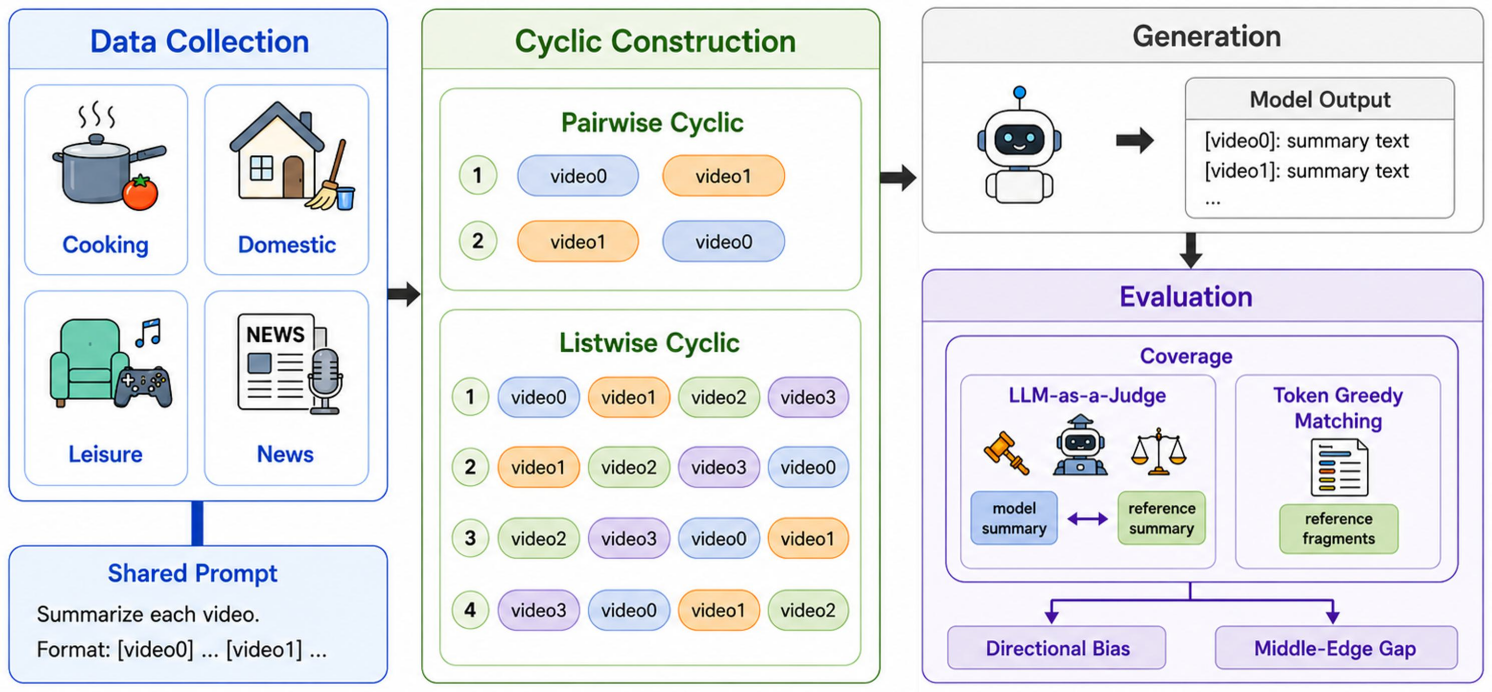}
\caption{Overview of our evaluation pipeline. We construct cyclic orderings for pairwise and listwise multi-video inputs, generate aligned per-video summaries, and evaluate positional effects with three complementary metrics.}
  \label{fig:framework}
\end{figure*}

Figure~\ref{fig:framework} illustrates our evaluation pipeline. We construct multi-video inputs, prompt the model to generate aligned per-video summaries, and evaluate positional effects with cyclic orderings across multiple settings, including model families, domains, video durations, and visual-budget variants. We then quantify positional effects
using three metrics: \textit{Coverage}, \textit{Directional Positional Bias}
(DPB), and \textit{Middle-Edge Gap} (MEG).

\subsection{Data Collection}
\label{sec:datasets}

We build our benchmark from two public datasets with human-written reference
summaries, covering four scenario categories: \textit{Cooking},
\textit{Domestic}, \textit{Leisure}, and \textit{News}. For \textit{News},
we use the News Video Dataset~\cite{whitehead-etal-2018-incorporating}. For the other
three categories, we sample from ActivityNet~\cite{caba2015activitynet}:
\textit{Cooking} corresponds to eating and cooking activities,
\textit{Domestic} to household activities, and \textit{Leisure} to social,
relaxation, and recreational activities.

These categories instantiate common online video-use settings. Prior work shows that users often encounter streams or sets of videos through recommendation-driven multi-video browsing, use online
video platforms for how-to learning, leisure viewing, and current events, and
increasingly consume news through social platforms
\cite{smith2018youtube,10.1145/2959100.2959190}.

To study the interaction between positional effects and video duration, we
partition inputs into short videos (0--1 minute) and longer-duration videos
(1--2 minutes). \textit{Cooking} and \textit{News} contain only short-video
settings, while \textit{Domestic} and \textit{Leisure} contain both short- and
longer-duration settings. For each domain--duration configuration, we randomly
sample 27 video groups. The four-video groups are non-overlapping within each
topic--duration configuration, while the pairwise and four-video settings may
share videos within the same topic. This yields 708 unique videos in total.
Although the number of unique videos is limited, cyclic-based evaluation
produces thousands of order-controlled instances, enabling systematic analysis
under a manageable inference budget.

\subsection{Summary Generation and Alignment}
\label{sec:summary_generation}

\textbf{Models.}
We evaluate both proprietary and open-source MLLMs. The main comparison includes InternVL3.5-8B, InternVL3.5-14B~\cite{wang2025internvl3}, Qwen3-VL-8B, Qwen3-VL-30B-A3B~\cite{bai2025qwen3vltechnicalreport}, MiniCPM-o-4.5~\cite{cui2026minicpmo45}, GLM-4.1V-9B-Thinking~\cite{glmvteam2025glm41vthinkingversatilemultimodalreasoning}, MiMo-VL-7B-RL~\cite{coreteam2025mimovltechnicalreport}, Gemini-3.1-Pro~\cite{gemini31pro_modelcard}, and GPT-5.4~\cite{openai2026gpt54}. 

\textbf{Input scale.}
We focus on two input scales: pairwise inputs with two videos ($P{=}2$) and listwise inputs with four videos ($P{=}4$). For each scale, we use a cyclic permutation design so that each video appears at every position exactly once. This supports balanced within-video positional analysis while keeping the number of permutations tractable. We use domain-scale notation (e.g., Cooking-2, News-4) to denote experiments on a specific domain with a given number of input videos.

\textbf{Input configuration.}
All main runs use the same task protocol: the task instruction and output
format are given first, followed by labeled video clips in cyclic order, with a
blank boundary frame between consecutive clips. We use \texttt{top\_p=1.0} and
temperature $0.9$ unless otherwise specified.

We set the requested summary length according to reference granularity:
ActivityNet-derived references average about 3.4 sentences per video, so we ask
for four sentences per video; News references average about 1.2 sentences, so
we ask for two sentences per video. For frame-based inputs, we uniformly sample frames,
using 16 frames per video for short-video and 24 frames for
longer ActivityNet settings when supported. The default resolution is
$448{\times}448$.

\textbf{Per-video summary alignment.}
We use a two-stage alignment strategy. First, we instruct the model to produce per-video summaries with a fixed template such as \texttt{[video0] ...}, \texttt{[video1] ...}, and extract the segments by pattern matching. When the output deviates from the template, we apply a fallback semantic alignment: we split the reference summaries and model output into sentences, compute sentence-level semantic similarity, and assign each generated sentence to the video with the highest similarity score. Sentences assigned to the same video are then concatenated into the aligned per-video summary.

\subsection{Evaluation Metrics}
\label{sec:metrics}

Let $P$ be the number of input videos and $p \in \{1,\dots,P\}$ the slot
position. We denote the coverage score of the video at position $p$ in instance
$i$ by $C_{i,p}$.

\subsubsection{Coverage.}
Coverage measures whether the generated summary preserves the key information
in the human-written reference. Because reference granularity differs across
datasets, we use dataset-adaptive estimators.

For ActivityNet-based domains, we use sentence-level semantic coverage. Given
reference sentences and the generated summary for the same video, GPT-5.1~\cite{openai2025gpt51addendum}
judges whether each reference sentence is covered, allowing paraphrases and
minor wording differences. Let $N_{\mathrm{cov}}^{(i,p)}$ be the number of
covered reference sentences and $N_{\mathrm{ref}}^{(i,p)}$ the total number of
reference sentences:
\begin{equation}
C_{i,p}
=
\frac{
N_{\mathrm{cov}}^{(i,p)}
}{
N_{\mathrm{ref}}^{(i,p)}
}.
\end{equation}

For News, references are much shorter, averaging about 1.2 sentences per video, and their key information often lies in specific named entities, locations, organizations, and event phrases. Binary sentence-level scoring is therefore too coarse: it can miss partial preservation of these surface-specific details. We instead use reference-fragment coverage inspired by NEWSROOM~\cite{grusky-etal-2018-newsroom}: we greedily match the longest contiguous reference spans appearing in the generated summary and compute the proportion of matched reference tokens:
\begin{equation}
C_{i,p}^{\mathrm{frag}}
=
\frac{
|\mathcal{M}_{i,p}|
}{
|r_{i,p}|
},
\end{equation}
where $\mathcal{M}_{i,p}$ is the set of matched reference-token positions and
$|r_{i,p}|$ is the number of reference tokens.

\subsubsection{Directional Positional Bias.}
Directional Positional Bias (DPB) measures earlier-versus-later preference. We
assign each position a linear weight
\begin{equation}
\alpha_p = \frac{2p - P - 1}{P - 1},
\end{equation}
where $\alpha_1=-1$ and $\alpha_P=1$. For instance $i$:
\begin{equation}
\mathrm{DPB}_i
=
\frac{
\sum_{p=1}^{P} \alpha_p C_{i,p}
}{
\sum_{p=1}^{P} C_{i,p} + \epsilon
},
\end{equation}
and aggregate over instances:
\begin{equation}
\mathrm{DPB}
=
\frac{1}{N}
\sum_{i=1}^{N}
\mathrm{DPB}_i.
\end{equation}
Negative values indicate earlier-position preference, and positive values
indicate later-position preference. Near-zero DPB does not imply position
invariance, since non-monotonic effects, such as middle-position weakness, can
cancel out in this signed measure. We therefore also use Middle-Edge Gap (MEG)
to capture whether middle positions underperform edge positions in four-video
inputs.

\subsubsection{Middle-Edge Gap.}
For four-video inputs ($P=4$), we define MEG as:
\begin{equation}
\mathrm{MEG}_i
=
\frac{C_{i,2}+C_{i,3}}{2}
-
\frac{C_{i,1}+C_{i,4}}{2},
\end{equation}
with aggregate
\begin{equation}
\mathrm{MEG}
=
\frac{1}{N}
\sum_{i=1}^{N}
\mathrm{MEG}_i.
\end{equation}
Negative MEG indicates middle-position weakness, while positive MEG indicates
a middle-position advantage.

\section{Experimental Results}
\label{sec:results}
\begin{table*}[t]
\centering
\small
\setlength{\tabcolsep}{3pt}
\resizebox{\textwidth}{!}{%
\begin{tabular}{llcccccccccccc}
\toprule
Model & Duration
& \multicolumn{1}{c}{Cooking-2}
& \multicolumn{2}{c}{Cooking-4}
& \multicolumn{1}{c}{Domestic-2}
& \multicolumn{2}{c}{Domestic-4}
& \multicolumn{1}{c}{Leisure-2}
& \multicolumn{2}{c}{Leisure-4}
& \multicolumn{1}{c}{News-2}
& \multicolumn{2}{c}{News-4} \\
\cmidrule(lr){3-3}
\cmidrule(lr){4-5}
\cmidrule(lr){6-6}
\cmidrule(lr){7-8}
\cmidrule(lr){9-9}
\cmidrule(lr){10-11}
\cmidrule(lr){12-12}
\cmidrule(lr){13-14}
 &  & DPB (\%) & DPB (\%) & MEG (\%) & DPB (\%) & DPB (\%) & MEG (\%) & DPB (\%) & DPB (\%) & MEG (\%) & DPB (\%) & DPB (\%) & MEG (\%) \\
\midrule
Qwen3-VL-8B & Short & -3.33 & -2.67 & -1.34 & 8.13 & -0.37 & -2.57 & 6.01 & 0.67 & -0.64 & 0.94 & 1.17 & 0.15 \\
Qwen3-VL-8B & Long & -- & -- & -- & -5.24 & -2.90 & -2.45 & 0.71 & -2.37 & -0.71 & -- & -- & -- \\
InternVL3.5-8B & Short & -3.34 & -4.16 & -3.34 & -3.27 & 1.67 & -4.50 & -0.15 & -2.07 & -0.10 & 1.67 & -0.99 & -1.04 \\
InternVL3.5-8B & Long & -- & -- & -- & 2.27 & -0.94 & -3.34 & 13.89 & -1.28 & -0.57 & -- & -- & -- \\
InternVL3.5-14B & Short & -5.93 & -1.74 & 0.40 & 4.06 & -2.51 & -2.11 & -3.04 & 0.07 & -2.65 & 1.61 & 0.92 & -1.49 \\
InternVL3.5-14B & Long & -- & -- & -- & -4.96 & -0.98 & -3.15 & -8.35 & -0.77 & -1.41 & -- & -- & -- \\
Qwen3-VL-30B-A3B & Short & -5.68 & -1.41 & 2.41 & 0.42 & 3.26 & -2.63 & 9.36 & -0.08 & -0.76 & 2.11 & 0.09 & -0.51 \\
Qwen3-VL-30B-A3B & Long & -- & -- & -- & 5.01 & -0.89 & -3.68 & -2.96 & 1.50 & -1.23 & -- & -- & -- \\
MiniCPM-o-4.5 & Short & -1.44 & -1.56 & -0.51 & -2.69 & 1.56 & 1.92 & 3.06 & 1.82 & 0.21 & -1.31 & -0.50 & -1.00 \\
MiniCPM-o-4.5 & Long & -- & -- & -- & -4.48 & 1.35 & -1.94 & -12.05 & -2.33 & -1.96 & -- & -- & -- \\
GLM-4.1V-9B-Thinking & Short & -0.78 & -2.06 & -2.35 & 0.05 & 0.40 & -1.22 & 4.87 & -1.10 & -2.82 & -0.89 & -0.72 & -1.46 \\
GLM-4.1V-9B-Thinking & Long & -- & -- & -- & -0.51 & 1.08 & 0.79 & -7.88 & -5.70 & -2.34 & -- & -- & -- \\
MiMo-VL-7B-RL & Short & 14.69 & -0.04 & -0.95 & -2.82 & 3.11 & 2.77 & 0.65 & 3.40 & -4.95 & 0.38 & 0.36 & -0.82 \\
MiMo-VL-7B-RL & Long & -- & -- & -- & -2.29 & 3.51 & 3.07 & 6.39 & 3.92 & 0.01 & -- & -- & -- \\
Gemini-3.1-Pro & Short & 2.54 & 3.72 & -0.73 & 17.88 & 2.48 & 0.47 & -1.81 & 2.05 & 0.05 & 0.05 & 0.60 & -0.18 \\
Gemini-3.1-Pro & Long & -- & -- & -- & -0.39 & 4.11 & 1.08 & 3.39 & 7.07 & -0.43 & -- & -- & -- \\
GPT-5.4 & Short & -1.79 & -3.15 & 1.52 & -14.97 & 1.86 & -0.78 & 6.49 & 3.22 & 1.20 & -0.29 & -0.86 & 0.53 \\
GPT-5.4 & Long & -- & -- & -- & 1.83 & 0.72 & 0.48 & 4.83 & 1.16 & -2.01 & -- & -- & -- \\
\bottomrule
\end{tabular}}
\caption{ Coverage-based Directional Positional Bias (DPB) and Middle-Edge Gap (MEG) across domains and models. Negative DPB values indicate primacy, i.e., higher coverage for earlier positions, while positive values indicate recency, i.e., higher coverage for later positions. MEG is defined only for four-video inputs; negative MEG values indicate middle-position weakness.}
\label{tab:base-bias-meg-matrix}
\end{table*}
\subsection{Main Results}
\label{sec:domain_effects}
Table~1 shows that positional effects do not reduce to a single universal primacy or recency pattern. Instead, the dominant pattern is heterogeneity across models and domains.
\paragraph{Middle-position weakness can be hidden by signed directional bias.}
DPB alone does not capture non-monotonic position effects. Several settings
with near-zero DPB still exhibit negative Middle-Edge Gap (MEG). For instance,
InternVL3.5-14B on Leisure-4 (Short) has DPB $0.07\%$ but MEG $-2.65\%$.
Similarly, GLM-4.1V-9B-Thinking has DPB $-1.10\%$ but
MEG $-2.82\%$.

\paragraph{Pairwise and listwise settings reveal different behavior.}
Pairwise directional bias does not reliably predict four-video behavior.
Several large $P{=}2$ effects shrink or reverse under $P{=}4$ inputs. For example,
Qwen3-VL-8B changes from a strong recency pattern on Domestic-2 (Short,
$+8.13\%$) to nearly zero DPB on Domestic-4 (Short, $-0.37\%$). InternVL3.5-8B
moves from strong recency on Leisure-2 (Long, $+13.89\%$) to mild primacy on
Leisure-4 (Long, $-1.28\%$). This suggests that listwise multi-video summarization introduces
position interactions that are not visible in pairwise evaluation.

\paragraph{Video duration changes the shape of positional effects.}
This can occur because longer-duration settings use a larger sampled-frame
budget, shifting the absolute input locations of later clips even when their
relative slot indices are unchanged. In Domestic-4, several models show more
negative MEG under longer inputs, such as Qwen3-VL-30B ($-2.63\%$ to
$-3.68\%$) and InternVL3.5-14B ($-2.11\%$ to $-3.15\%$). In Leisure-4,
negative MEG is common in both short and long settings, but some models change
substantially: GPT-5.4 moves from positive MEG in the short setting
($+1.20\%$) to negative MEG in the long setting ($-2.01\%$). These patterns suggest that duration and visual
context length can modulate positional effects, rather than simply amplifying a
fixed primacy or recency bias.

\paragraph{Case study.}
Among the proprietary models, we observe strong directional positional effects in both Gemini-3.1-Pro and GPT-5.4 on Domestic-2 (Short). GPT-5.4 has DPB
$-14.97\%$ on Domestic-2 (Short), indicating substantially lower coverage for
the second-position video. Manual inspection shows that the lower-scoring
second-position summaries are often more generic, omitting concrete actions or object details. For example, summaries describing workers
near a concrete or plaster machine were not judged as covering the reference
phrase about men making plaster, and summaries mentioning a pit stop or loose
tires were not judged as covering the specific action of changing tires.

This case also shows that small DPB values should be interpreted carefully.
Near-zero DPB may reflect balanced coverage, but it may also occur when a model
is uniformly generic and fails to cover all videos well. Stronger models may
therefore exhibit larger measured positional bias because they preserve more
details in favored positions, making uneven coverage easier to detect.

We also find that coverage differences can be accompanied by uneven output allocation. Table~\ref{tab:qwen8b-domestic-short-allocation} shows Qwen3-VL-8B on Domestic-4 (Short). The middle slots have comparable word counts to the first slot but lower coverage, while the final slot is longer and more often exceeds the requested number of sentences.

\begin{table}[t]
\centering
\small
\setlength{\tabcolsep}{5pt}
\begin{tabular}{lccc}
\toprule
Position & Avg. words & Over-rate & Coverage \\
\midrule
Pos. 1 & 66.92 & 2.78\% & 0.4388 \\
Pos. 2 & 66.60 & 1.85\% & 0.4118 \\
Pos. 3 & 67.09 & 6.48\% & 0.4047 \\
Pos. 4 & 69.68 & 8.33\% & 0.4290 \\
\bottomrule
\end{tabular}
\caption{\textbf{Output allocation example for Qwen3-VL-8B on Domestic-4 (Short).}
Over-rate denotes the proportion of outputs exceeding the requested number of
sentences.}
\label{tab:qwen8b-domestic-short-allocation}
\end{table}
\subsection{Visual-budget robustness.}
We further examine whether the observed positional effect is an artifact of
a particular visual input configuration. This analysis uses Qwen3-VL-8B on the
$P{=}4$ short-video setting. We vary the number of sampled frames while fixing
the resolution at $448{\times}448$, and vary the resolution while fixing the
frame count at 16. We report
MEG (\%), where negative values indicate that middle positions underperform
edge positions.

\begin{figure}[t]
    \centering
    \includegraphics[width=\linewidth]{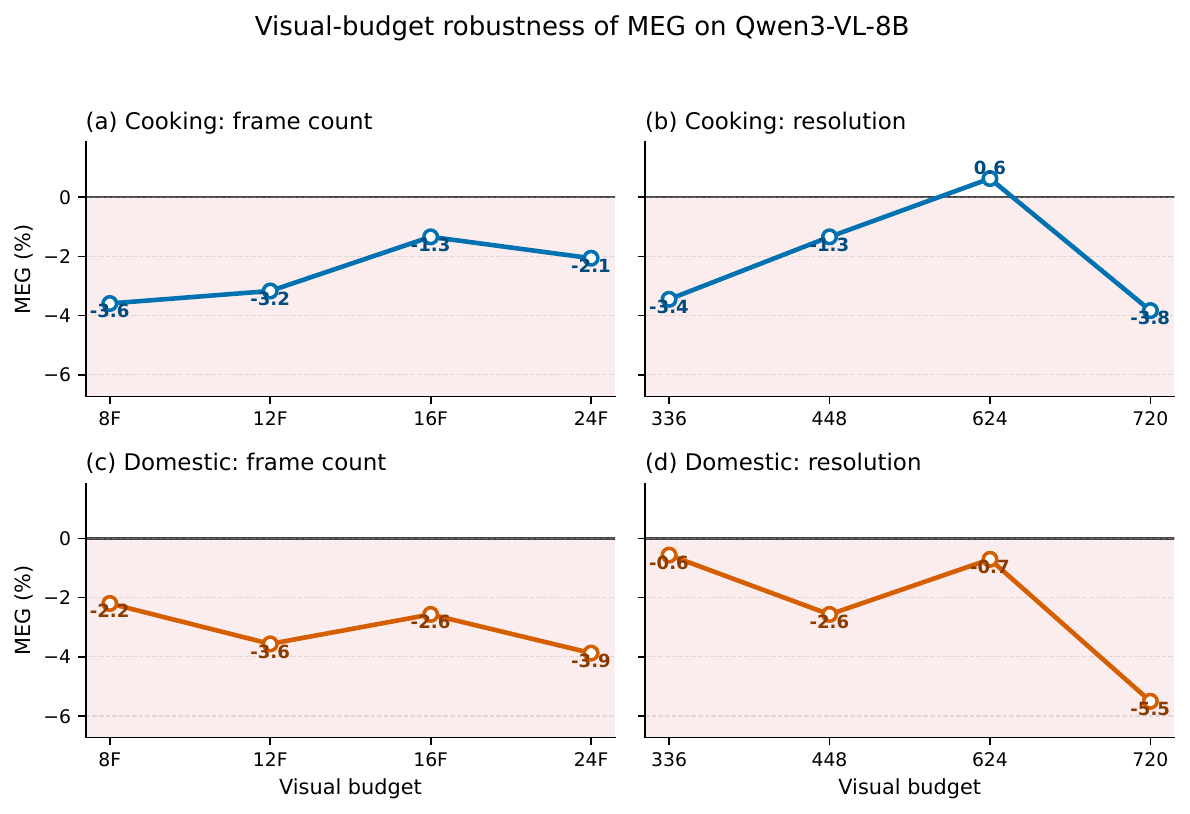}
    \caption{
    Visual-budget robustness check for Qwen3-VL-8B.
    }
    \label{fig:qwen8_visual_budget_meg}
\end{figure}

Figure~\ref{fig:qwen8_visual_budget_meg} shows that the negative MEG pattern is
largely preserved under visual-budget changes. In
\textit{Domestic}, MEG remains negative under all tested settings. 

We also run the same visual-budget ablation on MiMo-VL-7B-RL, whose more stable effect appears in Coverage: using only 8 frames reduces average Coverage across domains, while 24 frames improves it. We report the full MiMo ablation in Appendix~\ref{app:visual_budget_tables}.
\subsection{Requested Summary Budget and Middle-Position Weakness}
\label{sec:prompt_budget}

We next test whether middle-position weakness is driven by the summary-length constraint in the prompt. 
A tight requested budget may force the model to omit details unevenly across positions. 
We therefore vary the requested per-video budget with three settings: short, base, and long, corresponding to 1/2, 2/4, and 3/6 requested sentences for News/ActivityNet-based domains, respectively. 
For each four-video input, we compute MEG and also report an exact-budget subset whose generated sentence counts match the requested numbers.
\begin{table}[t]
\centering
\small
\setlength{\tabcolsep}{5pt}
\begin{tabular}{lccc}
\toprule
Budget & Requested sent. & All outputs & Exact-budget \\
\midrule
Short & 2 & $-2.29$ & $-0.63$ \\
Base  & 4 & $-2.45$ & $-2.60$ \\
Long  & 6 & $-4.22$ & $-3.70$ \\
\bottomrule
\end{tabular}
\caption{
MEG for the Domestic-long setting under different requested summary budgets
}
\label{tab:domestic_long_prompt_budget}
\end{table}

Table~\ref{tab:domestic_long_prompt_budget} illustrates the pattern on Domestic-long. Under the short budget, the gap becomes much smaller in the exact-budget subset, suggesting that a tight two-sentence budget compresses all positions and can mask position-specific differences. With larger budgets, the gap remains negative even after sentence-count matching, and is largest in the long setting. Thus, giving the model more output space does not necessarily remove middle-position weakness; it can instead make uneven detail allocation more visible.

More broadly, increasing the requested budget does not benefit all positions equally. From short to long, middle positions receive smaller coverage gains than edge positions in four of six \(P=4\) datasets. This suggests that the issue is not only whether the model writes enough sentences, but how added generation space is allocated across input positions.

\subsection{Boundary-format Robustness}
\label{sec:boundary_format}

Prior multi-video benchmarks show that MLLMs can struggle to filter irrelevant information across videos in multi-video inputs~\cite{peng2025mvueval}. This makes input boundary design a relevant protocol factor: visual separators may help models segment videos, while text-only labels may change how strongly the model separates one video from another. We therefore compare the default black-frame separator with a text-label-only variant.

Overall, changing the boundary format affects the position-wise coverage curve, but does not provide a consistent improvement. For InternVL3.5-8B, the text-label-only format changes the curve more visibly: \textit{Cooking} and \textit{Domestic} show less negative MEG, but this does not translate into uniformly better summary quality. In \textit{Domestic}, position 2 improves while position 4 drops substantially, suggesting that text-only boundaries flatten the curve rather than consistently improving coverage. For Qwen3-VL-8B, removing black frames does not remove the middle-position weakness.

\begin{figure}[t]
    \centering
    \includegraphics[width=\linewidth]{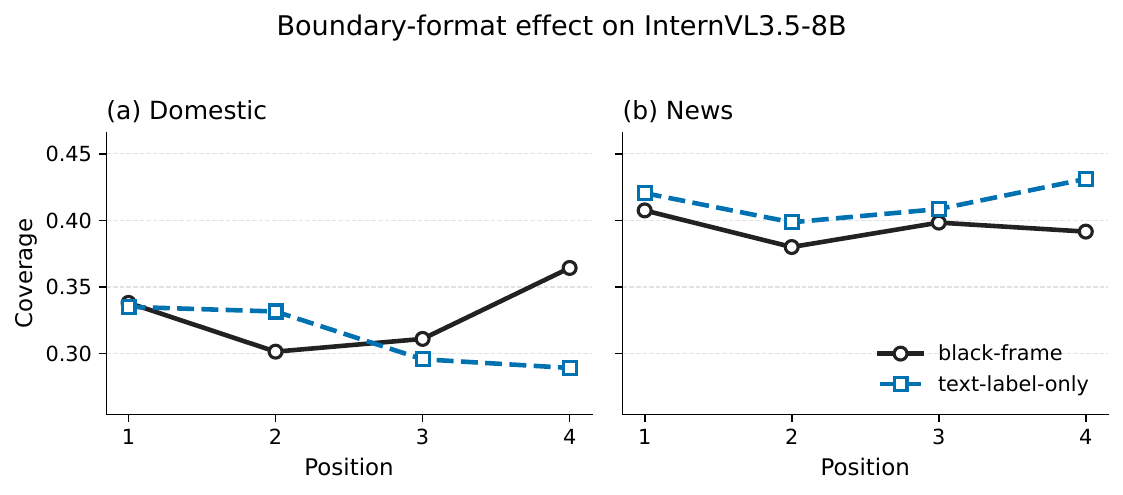}
    \caption{
    Boundary-format comparison for InternVL3.5-8B.
    }
    \label{fig:intern8_boundary_format}
\end{figure}

As a slot-boundary sanity check, we also compare each generated slot summary with its own reference and with the other references in the same group using TF-IDF content-token similarity. A slot is marked as potential contamination only when it is substantially more similar to another video's reference than to its own. This proxy is not a verified leakage label, but it can reveal large-scale slot-boundary failures. We do not observe a systematic increase in potential contamination under text-label-only formatting.
\section{Prompt-Level Mitigation and Diagnostics}
\label{sec:mitigation}
We explore lightweight prompt-level interventions and diagnostics that do not require model retraining or architectural access. 
We first test an explicit balanced-attention instruction, then evaluate a single-target generation protocol, and finally examine prompt placement as a target-binding diagnostic.

\subsection{Balanced-attention instruction.}
\label{sec:prompt_mitigation}

\begin{figure}[t]
    \centering
    \includegraphics[width=\linewidth]{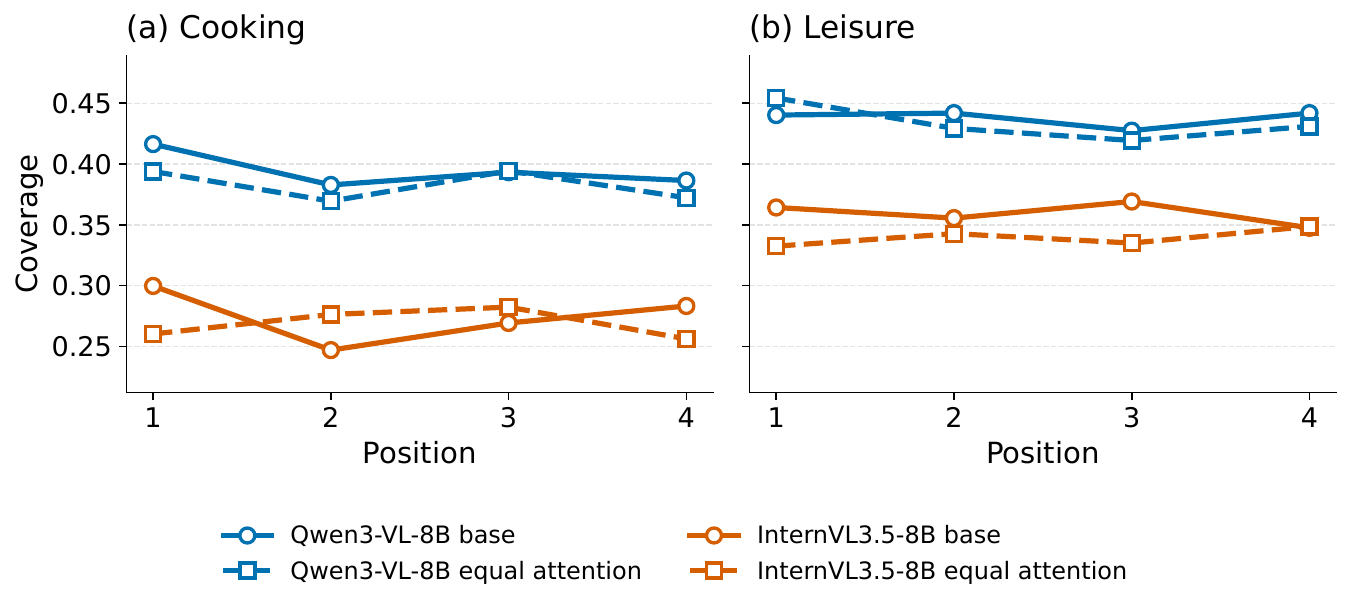}
    \caption{
    Effect of equal-attention prompting on position-wise coverage for
    Qwen3-VL-8B and InternVL3.5-8B.
    }
    \label{fig:equal_attention_curves}
\end{figure}

We first add a direct instruction asking the model to allocate equal attention
to each video. Figure~\ref{fig:equal_attention_curves}
shows that this intervention does not provide reliable mitigation. For
Qwen3-VL-8B, the equal-attention prompt does not produce a stable coverage
balancing effect across positions. For InternVL3.5-8B, the Cooking curve becomes
flatter, but this flattening partly comes from lowering edge-position coverage
rather than uniformly improving all positions. Thus, the prompt can reshape the
position-wise coverage curve, but does not consistently improve summary quality
or remove positional imbalance.

\subsection{Single-target Prompting}
\label{sec:single_target_prompting}

We also test a single-target generation protocol as an output-side diagnostic. The model receives the same multi-video input, but each call requests only one specified target-video summary instead of all summaries in one response. This removes previously generated summaries from the autoregressive prefix and tests whether position-dependent errors arise from joint generation effects, such as uneven detail allocation, or cross-summary contamination. If the bias persists, it is more likely due to the multi-video input itself; if it weakens, joint multi-summary generation may contribute to the effect.

\paragraph{Target-addressing concern.}
Single-target prompting introduces an additional target-addressing ambiguity that is less salient in joint per-video generation. Unlike joint generation, where the model outputs descriptions for all videos together, single-target prompting requires the model to resolve one specified label. We observe that models may confuse this label with ordinal position: even when the prompt explicitly states which frames correspond to each label, \texttt{video[1]} can still be treated as the first or currently queried video rather than the second zero-indexed video. 

To quantify this label-resolution issue, we compute TF-IDF cosine similarity
between each generated summary and the four references in the same group.
A target hit occurs when the true target reference has the highest similarity; a margin confusion occurs when another reference
exceeds the true target by over 0.05. Table~\ref{tab:single_target_indexing}
shows that one-indexed labels mainly improve target addressing at position~2,
where zero-indexing is most ambiguous. We therefore compare zero-indexed and one-indexed labels on InternVL3.5-14B to test whether this addressing convention contributes to the single-target effect.

\begin{table}[t]
\centering
\small
\setlength{\tabcolsep}{5pt}
\resizebox{\linewidth}{!}{%
\begin{tabular}{lccc}
\toprule
Target pos. & $\Delta$Coverage & $\Delta$Target hit & $\Delta$Margin confusion \\
\midrule
Pos. 1 & $-0.54$ & $-0.23$ & $-0.93$ \\
Pos. 2 & $+4.18$ & $+14.12$ & $-11.81$ \\
\bottomrule
\end{tabular}
}
\caption{
Effect of changing single-target labels from zero-indexed to one-indexed form
on InternVL3.5-14B.
}
\label{tab:single_target_indexing}
\end{table}

\begin{table}[t]
\centering
\small
\setlength{\tabcolsep}{4pt}
\renewcommand{\arraystretch}{1.05}
\resizebox{\linewidth}{!}{%
\begin{tabular}{lrrrr}
\toprule
Scale & $\Delta$Weak Cov. & $\Delta$Weak Sent. & Weak pos. imp. & $\Delta$Density \\
\midrule
$P{=}2$ & $-4.09$ & $+0.16$ & $0/6$ & $-0.121$ \\
$P{=}4$ & $+2.87$ & $+1.26$ & $4/6$ & $-0.224$ \\
\bottomrule
\end{tabular}
}
\caption{
Effect of single-target prompting relative to joint generation for InternVL3.5-14B.
 Weak positions are identified from the joint-generation setting. $\Delta$Density is the change in coverage per 100 words.
}
\label{tab:single_target_scale_summary}
\end{table}

\paragraph{Effect analysis.}
Table~\ref{tab:single_target_scale_summary} shows a scale-dependent effect. For $P{=}4$, single-target prompting improves the weak-position coverage by 2.87 percentage points and improves the weakest joint-generation position in 4 of 6 datasets. The gain is accompanied by a much larger sentence increase, especially for positions that were compressed in the joint output.

In contrast, for $P{=}2$, the weak-position coverage decreases despite a small increase in generated sentences. Overall, this protocol is not a general summarization improvement, but a possible inference-time workaround when the number of input videos grows and joint generation begins to compress some summary slots.

\subsection{Prompt Placement and Leakage Audit}
\label{sec:prompt_placement}
We next examine whether prompt placement affects target binding. In the main setting,
the task instruction appears before the videos, while a video-first variant
places the same instruction after the videos. We first inspect the easier $P{=}2$ setting, where inputs
are shorter and model performance is generally stronger. For InternVL3.5-14B,
video-first produces slightly longer ActivityNet outputs on average
(3.817 vs. 3.667 sentences), but reduces $P{=}2$ coverage by 4.29 points. This suggests that prompt placement affects target binding, not merely instruction compliance.

To further test this target-binding concern, we conduct a targeted leakage audit. A full target-alignment audit would
require comparing thousands of generated summaries with their corresponding videos. Instead, we use a reference-based
candidate miner to reduce the review space. The miner is designed around the
observation that cross-video leakage often appears as a concrete action or
object detail from another video, even when clips have similar scenes. For each
generated sentence, it compares concrete action/object anchor words against the
reference of the claimed target video and the references of the other videos in
the same group. A sentence is selected as a candidate when its concrete evidence
is weakly supported by the claimed target but better supported by another video.
We then manually inspect these candidates against the videos and count a
verified leak only when the sentence describes another input video.

\begin{table}[t]
\centering
\small
\setlength{\tabcolsep}{5pt}
\resizebox{\linewidth}{!}{%
\begin{tabular}{lccccc}
\toprule
Protocol & Cand. sent. & Verified leak & Cand. prec. & Slot rate & Sent. rate \\
\midrule
Prompt-first & 263 & 52 & 19.77\% & 1.93\% & 0.68\% \\
Video-first  & 446 & 96 & 21.52\% & 3.56\% & 0.93\% \\
\bottomrule
\end{tabular}
}
\caption{
Manual verification of candidate cross-video leakage under prompt-placement variants.
Cand. sent. denotes mined candidate sentences; Verified leak denotes manually confirmed leakage sentences; Cand. prec. is Verified leak / Cand. sent.
}
\label{tab:prompt_placement_leakage_rates}
\end{table}

Table~\ref{tab:prompt_placement_leakage_rates} shows that video-first prompting
produces more mined candidates and more manually verified leakage sentences
than prompt-first prompting. At the position level, candidate precision is
higher for positions 2 and 3 than for positions 1 and 4 under both prompt
placements. We observe two recurring failure modes: full target-addressing
failures, where the summary is assigned to the wrong video, and local detail
intrusions, where the summary mostly describes the target video but borrows a
concrete action or object from another video. Since one target-addressing error
can affect multiple generated sentences, sentence-level counts may overstate
the number of independent leakage events. 
\section{Conclusion}
\label{sec:conclusion}

We present a systematic study of positional bias in multi-video summarization for MLLMs. Our results show that current models are not reliably position-invariant: summary quality varies with input order across domains, model families, and input settings. We further show that signed directional bias alone is insufficient, since low DPB can still hide middle-position weakness under MEG. Robustness and protocol analyses suggest that these effects are not reducible to a single artifact, while lightweight prompt-level interventions only partially reshape them. Our benchmark, metrics, and diagnostics provide a foundation for studying and improving order robustness in multi-video summarization.
\section{Limitations}

\textbf{Input scale and domains.}
Our main experiments focus on inputs of up to four videos and clips within the 0--2 minute range from selected scenario categories. Positional effects
under longer video lists, longer clips, or more open-ended video mixtures remain
open questions.

\textbf{Evaluation scope.}
Our coverage scores depend on human-written references. ActivityNet-derived
domains and News also use different coverage estimators because their reference
granularities differ, so absolute scores should be compared within each dataset
family rather than across families. Our leakage audits are diagnostic and
candidate-based, not exhaustive annotations of all generated content.

\textbf{Mechanistic and intervention scope.}
We mainly provide behavioral and protocol-level analyses. Although our
ablation and mitigation experiments identify factors that reshape positional
effects, they do not provide a full causal mechanism. Also, stronger mitigation may
require training-time or architectural interventions beyond lightweight
prompting.
\section{Potential Risks}

A potential risk of this work is that positional bias may be misused to place preferred videos in fixed positions, thereby shaping summary emphasis and causing overexposure of some content while reducing the visibility of others.

\bibliography{custom}
\clearpage
\appendix

\section{Prompts and LLM as a judge}
\label{sec:appendix}

\subsection{Per-video Summarization Prompts}

We use a fixed per-video output structure to support reliable slot extraction and position-wise analysis. In the main experiments, the task prompt is placed before the visual inputs. For ActivityNet-based domains, we request four sentences per video. For the News domain, whose references are much shorter, we request two sentences per news video. The required output markers are zero-indexed and extend to the number of videos in the input, e.g., \texttt{[video0]} through \texttt{[video3]} for four-video inputs.

\begin{tcolorbox}[
  title={Main black-frame prompt.},
  colback=gray!10,
  colframe=black!50,
  colbacktitle=gray!20,
  coltitle=black,
  fonttitle=\bfseries,
  boxrule=0.5pt,
  arc=1mm,
  left=1mm,
  right=1mm,
  top=1mm,
  bottom=1mm
]
\small
\ttfamily
Different videos in the input are separated by black frames.\\
Summarize each video in 4 sentences.\\
Use exactly this format, including the square brackets:\\
\lbrack{}video0\rbrack{} summary\\
\lbrack{}video1\rbrack{} summary\\
...\\
\lbrack{}video\{P-1\}\rbrack{} summary\\[1mm]
\textnormal{\emph{Input layout:}}\\
Video 0: \{N\} uniformly sampled frames\\
\{frames of video 0\}\\
\textless{}black separator\textgreater{}\\
Video 1: \{N\} uniformly sampled frames\\
\{frames of video 1\}\\
\textless{}black separator\textgreater{}\\
...\\
Video \{P-1\}: \{N\} uniformly sampled frames\\
\{frames of video \{P-1\}\}
\end{tcolorbox}

\begin{tcolorbox}[
  title={Text-label-only prompt.},
  colback=gray!10,
  colframe=black!50,
  colbacktitle=gray!20,
  coltitle=black,
  fonttitle=\bfseries,
  boxrule=0.5pt,
  arc=1mm,
  left=1mm,
  right=1mm,
  top=1mm,
  bottom=1mm
]
\small
\ttfamily
Different videos in the input are separated by text labels only.\\
Summarize each video in 4 sentences.\\
Use exactly this format, including the square brackets:\\
\lbrack{}video0\rbrack{} summary\\
\lbrack{}video1\rbrack{} summary\\
...\\
\lbrack{}video\{P-1\}\rbrack{} summary\\[1mm]
\textnormal{\emph{Input layout:}}\\
Video 0: \{N\} uniformly sampled frames\\
\{frames of video 0\}\\
--- Text separator: end of Video 0; next is Video 1. ---\\
Video 1: \{N\} uniformly sampled frames\\
\{frames of video 1\}\\
--- Text separator: end of Video 1; next is Video 2. ---\\
...\\
Video \{P-1\}: \{N\} uniformly sampled frames\\
\{frames of video \{P-1\}\}
\end{tcolorbox}

For News, the second line is again replaced by \texttt{Summarize each news video in 2 sentences.}

\subsection{LLM-as-a-Judge Prompt for Sentence-level Coverage}

We evaluate ActivityNet Coverage using an LLM-as-a-Judge protocol with \texttt{gpt-5.1-2025-11-13}. Given a reference summary split into sentences and a model-generated summary for the same video, the judge returns a binary covered/not-covered decision for each reference sentence. We use the following prompt with temperature 0.

\begin{tcolorbox}[colback=gray!10, colframe=black!50, boxrule=0.5pt, arc=1mm, left=1mm, right=1mm, top=1mm, bottom=1mm]
\small
\ttfamily
You are evaluating whether a model-generated video summary covers a human reference summary.\\
For EACH reference sentence, decide whether it is clearly covered by the model-generated summary.\\
\\
Scoring rule:\\
- covered = 1 if the model summary explicitly or implicitly describes the same fact as the reference sentence.\\
- covered = 0 if the model summary omits the key information, contradicts the reference sentence, or is too vague/general to support the specific reference sentence.\\
\\
Rules:\\
- Different wording is acceptable when the meaning matches.\\
- Do not require exact phrasing.\\
- If only part of a reference sentence is covered and an important fact is missing, score 0.\\
- For covered=0, explain briefly what is missing or contradicted.\\
- For covered=1, reason must be an empty string.\\
\\
Reference sentences:\\
\{numbered\_refs\}\\
\\
Model-generated summary:\\
\{model\_summary\}\\
\\
Return ONLY valid JSON. Do not include markdown.\\
Required JSON schema:\\
\{\\
~~"sentence\_scores": [\\
~~~~\{"index": 1, "covered": 1, "reason": ""\},\\
~~~~\{"index": 2, "covered": 0, "reason": "The model summary does not mention ..."\}\\
~~]\\
\}
\end{tcolorbox}

\subsection{Human Agreement and Judge Robustness}
\label{sec:appendix_judge_robustness}

To assess the reliability of the sentence-level coverage judge, we recruited three undergraduate annotators and compensated them at a rate of USD 8 per hour to conduct a human agreement check on 240 annotated items. 
Each annotator independently scored coverage, and we compare pairwise human agreement with the agreement between the LLM judge and the mean human score. 
Table~\ref{tab:human_llm_judge_agreement} reports Pearson and Spearman correlations.

\begin{table}[t]
\centering
\small
\setlength{\tabcolsep}{5pt}
\resizebox{\linewidth}{!}{%
\begin{tabular}{lccc}
\toprule
Comparison & $n$ & Pearson & Spearman \\
\midrule
Annotator 1 vs. Annotator 2 & 240 & 0.7760 & 0.7651 \\
Annotator 1 vs. Annotator 3 & 240 & 0.7152 & 0.7138 \\
Annotator 2 vs. Annotator 3 & 240 & 0.7765 & 0.7796 \\
Human mean vs. LLM & 240 & 0.7116 & 0.6985 \\
\bottomrule
\end{tabular}
}
\caption{
Human agreement and LLM-judge robustness for sentence-level coverage scoring.
The LLM judge shows a positive correlation with the mean human score.
}
\label{tab:human_llm_judge_agreement}
\end{table}

The LLM judge is not a perfect substitute for human annotation, but its
agreement with the mean human score is comparable to the lower end of
human-human agreement. We therefore use it as a scalable coverage estimator for
large-scale positional analysis, while treating individual borderline cases with
caution.

\section{Dataset Licenses and Usage Conditions}
\label{app:data_governance}

Our benchmark is constructed from two publicly available third-party datasets derived from web videos: the News Video Dataset and ActivityNet. For the News Video Dataset, the official repository states that the dataset is released for educational and research purposes only, while the videos remain protected under their respective YouTube licenses and belong to their original copyright holders. For ActivityNet, we rely on the official release of the dataset and the access channels provided by the benchmark creators.

\section{More Experimental Details}
\paragraph{Computational budget.}
All local experiments were conducted on NVIDIA RTX PRO 6000 GPUs with 96GB
memory. Qwen3-VL-30B-A3B and InternVL3.5-14B were run with three GPUs, while
the other locally evaluated models were run with a single GPU. Across local
inference and automatic evaluation runs, the total computational budget was
approximately 1,100 GPU-hours.

\subsection{Parsing and Output-Failure Diagnostics}
\label{app:parsing_failure}

We also inspect whether positional effects are confounded by parsing or
complete output failures. Most models follow the requested \texttt{[videoX]}
format. We distinguish two failure modes. A \textit{marker failure} occurs when
the model attempts the fixed marker format but misses or malforms one or more
required video slots; such cases are marked as failures for the affected slots.
A \textit{fallback} case occurs only when the model does not follow the fixed
marker format at all, in which case we use BERTScore sentence assignment to
recover a best-effort alignment. Table~\ref{tab:marker_fallback_frequency}
reports only failure counts.

\begin{table}[t]
\centering
\small
\setlength{\tabcolsep}{4pt}
\begin{tabular}{lrr}
\toprule
Model & Marker fail. & Fallback \\
\midrule
Qwen3-VL-8B & 1 & 1 \\
InternVL3.5-8B & 5 & 1 \\
InternVL3.5-14B & 8 & 0 \\
Qwen3-VL-30B-A3B & 1 & 0 \\
MiMo-VL-7B-RL & 14 & 3 \\
\bottomrule
\end{tabular}
\caption{
Marker parsing failures in the baseline records.
}
\label{tab:marker_fallback_frequency}
\end{table}

Complete output failures are more concentrated. Table~\ref{tab:mimo_failure_settings}
shows the MiMo-VL-7B-RL settings where the model produced placeholder-only
outputs, such as literal \texttt{[video0] summary}-style text, rather than
substantive descriptions of the videos. These outputs are unusable for coverage
evaluation and are assigned zero Coverage in the baseline aggregation. These failures occur mainly in ActivityNet settings and become more
frequent in four-video inputs.

\begin{table}[t]
\centering
\small
\setlength{\tabcolsep}{6pt}
\begin{tabular}{lr}
\toprule
Setting & Failures \\
\midrule
Cooking-2 & 3 \\
Cooking-4 & 22 \\
Domestic-S-2 & 6 \\
Domestic-S-4 & 24 \\
Domestic-L-2 & 7 \\
Domestic-L-4 & 22 \\
Leisure-S-2 & 10 \\
Leisure-S-4 & 28 \\
Leisure-L-2 & 8 \\
Leisure-L-4 & 36 \\
\bottomrule
\end{tabular}
\caption{
MiMo-VL-7B-RL complete output failures.
}
\label{tab:mimo_failure_settings}
\end{table}

\subsection{Coverage Curves for All Baseline Models}
\label{app:all_position_curves}

Figures~\ref{fig:base_position_curves_group_a} and
\ref{fig:base_position_curves_group_b} report the full baseline
Coverage--position curves. We split models into two groups only for readability;
both figures use the same ordering of domains, durations, and input scales.

Some Gemini-3.1-Pro settings have lower Coverage despite fluent summaries.
This reflects the intended scope of our metric rather than a global judgment of
summary quality.Coverage is reference-grounded: since summaries are ultimately intended for human readers, we assume that human-written references identify the salient facts and events that readers expect a summary to preserve. Our
inspection suggests that Gemini-3.1-Pro often produces shorter and more
abstractive summaries, which may contain fewer explicit reference-grounded
details under this metric. This limitation does not undermine the positional
analysis, because we compare Coverage across input slots under the same scoring
rule. If reference-grounded information preservation varies by slot, the model
still exhibits order-dependent summarization behavior.
\newcommand{\curvepanel}[2]{%
\begin{minipage}[t]{0.32\textwidth}
\centering
\includegraphics[width=\linewidth]{#1}
\vspace{-1.5mm}
\caption*{\scriptsize #2}
\end{minipage}%
}

\newcommand{\curvelegenditem}[3]{%
\textcolor[HTML]{#1}{\rule{1.2em}{1.3pt}}\hspace{0.4mm}#3%
}

\begin{figure*}[p]
\centering

\curvepanel{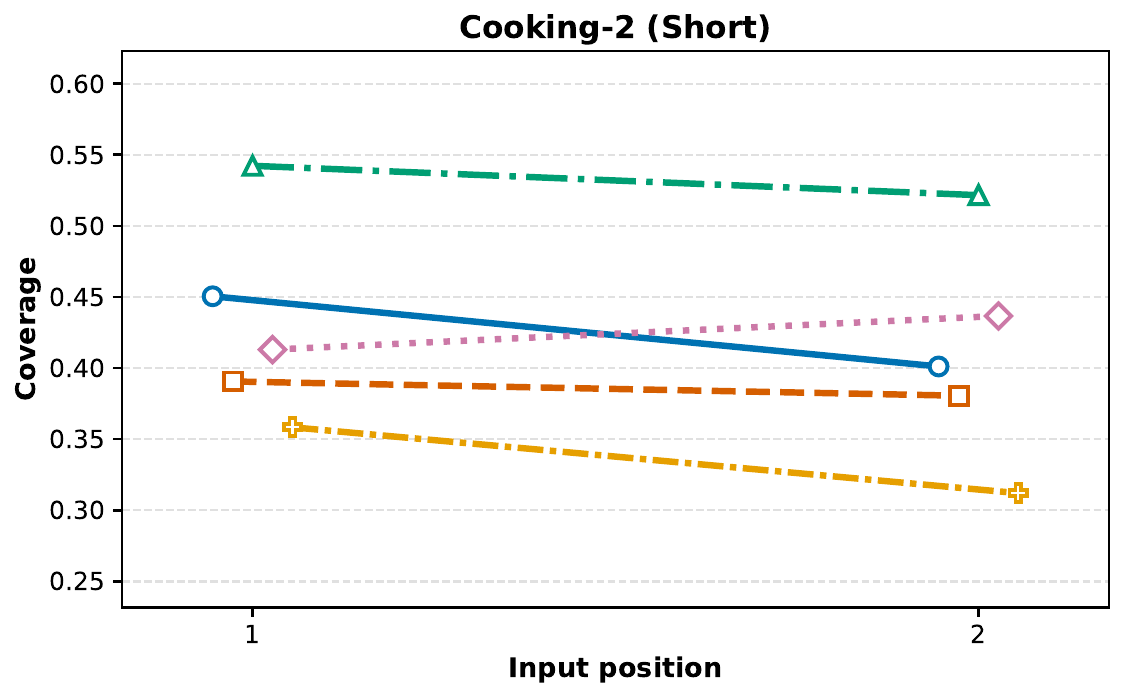}{Cooking-2 Short}
\hfill
\curvepanel{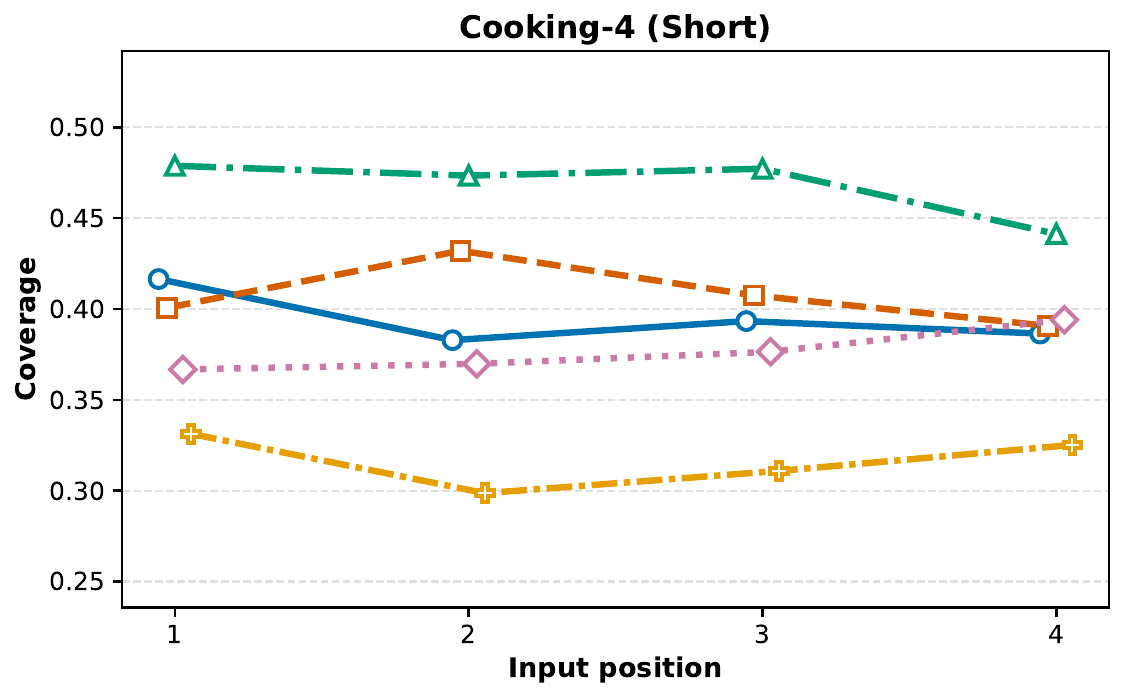}{Cooking-4 Short}
\hfill
\curvepanel{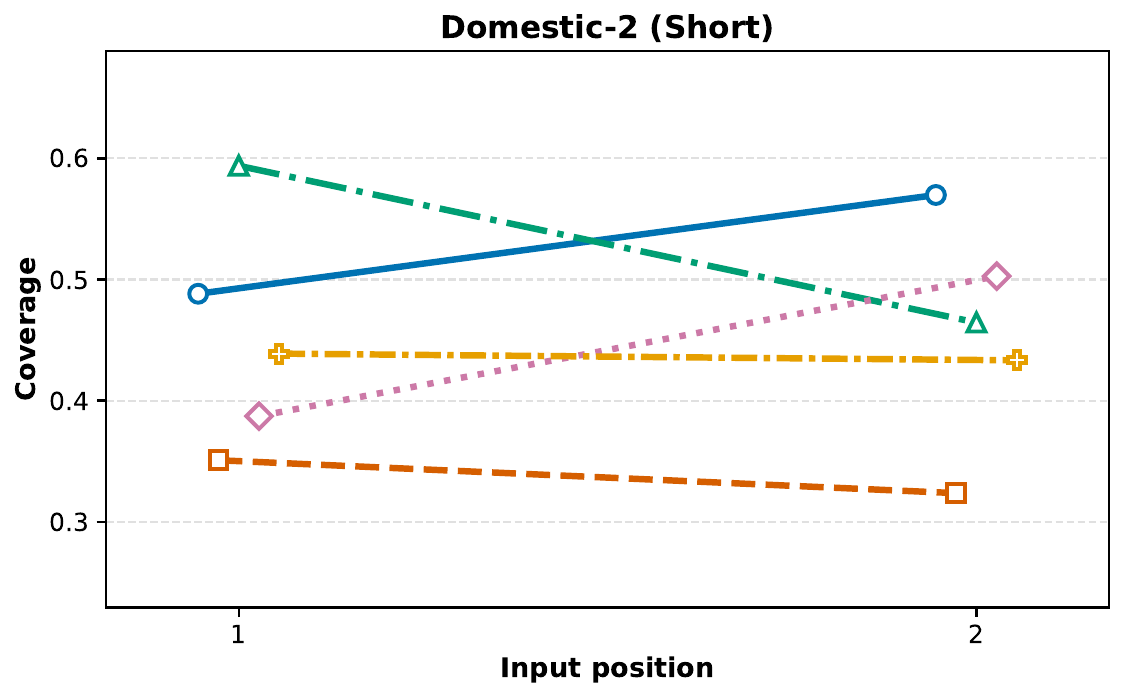}{Domestic-2 Short}

\vspace{1mm}

\curvepanel{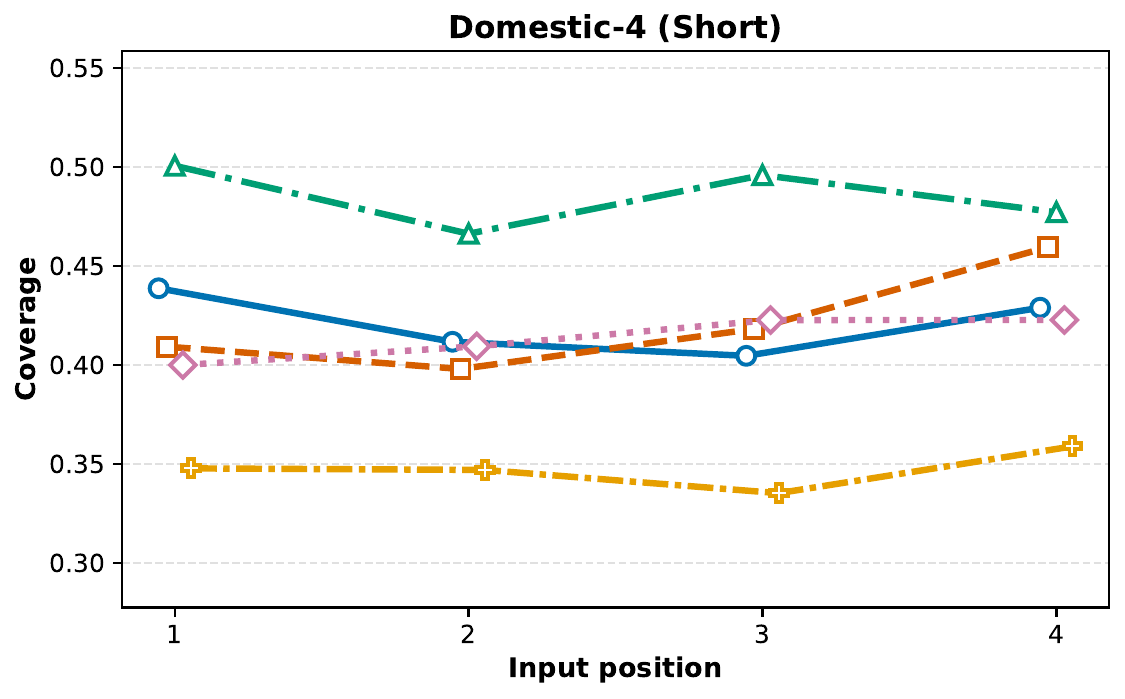}{Domestic-4 Short}
\hfill
\curvepanel{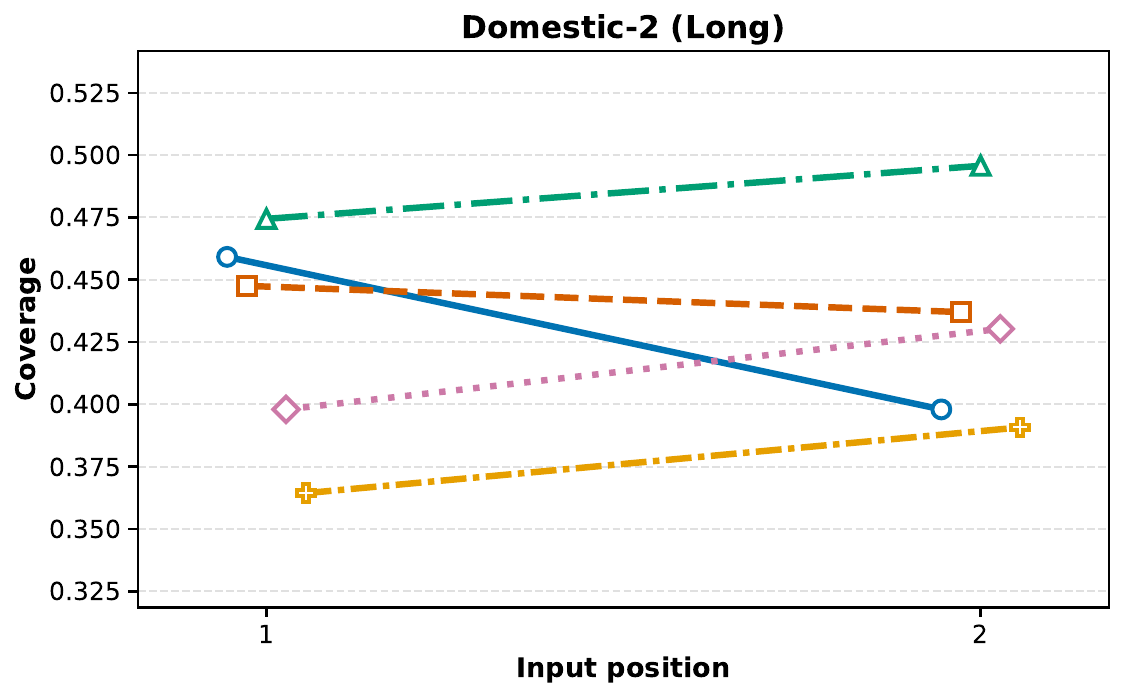}{Domestic-2 Long}
\hfill
\curvepanel{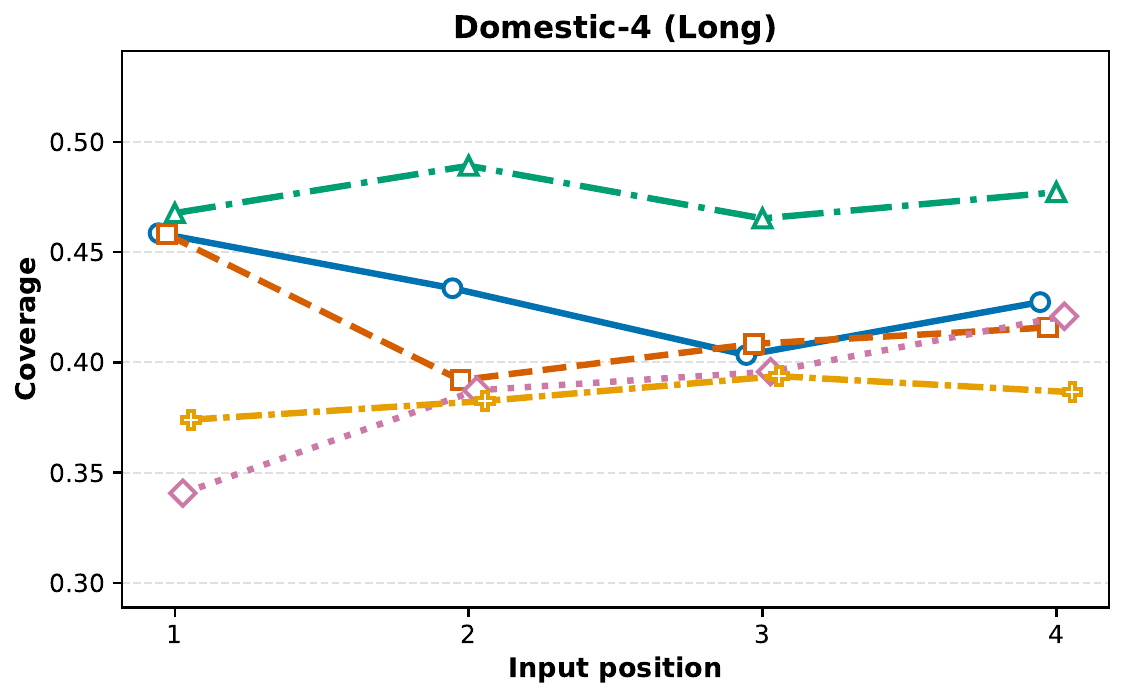}{Domestic-4 Long}

\vspace{1mm}

\curvepanel{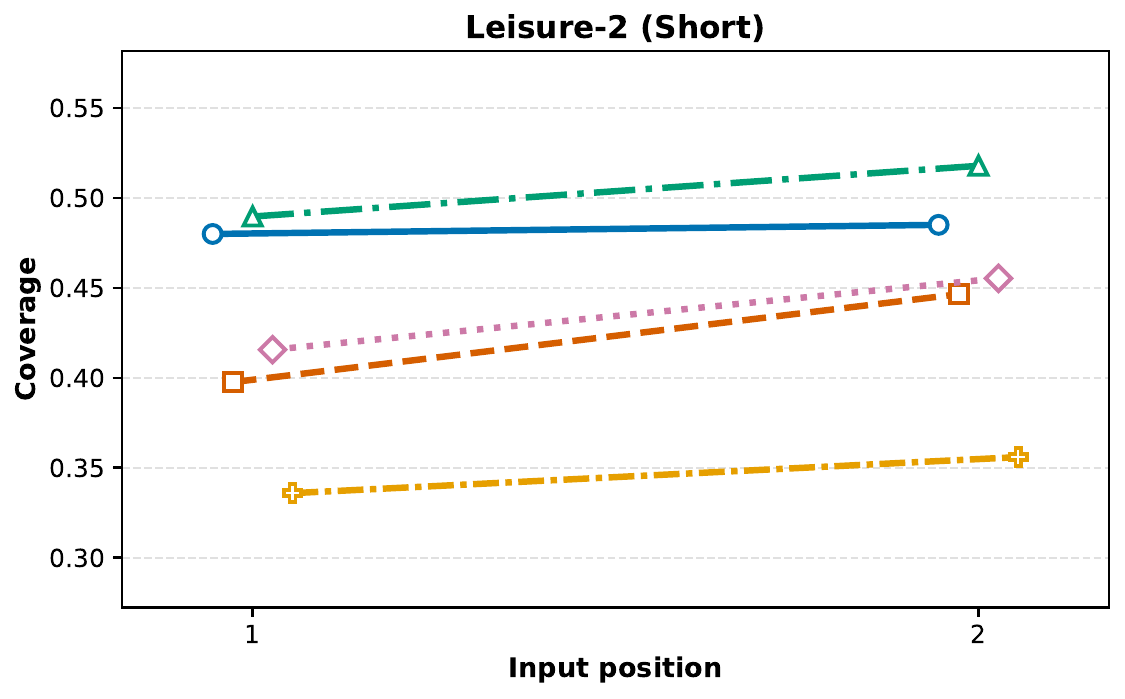}{Leisure-2 Short}
\hfill
\curvepanel{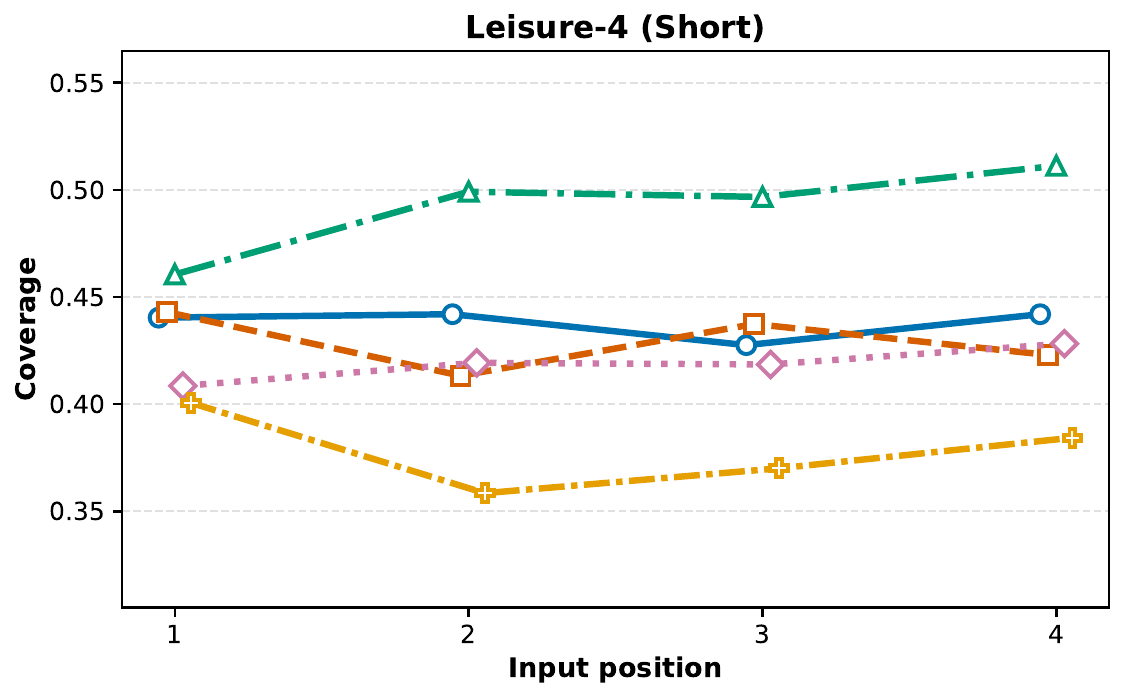}{Leisure-4 Short}
\hfill
\curvepanel{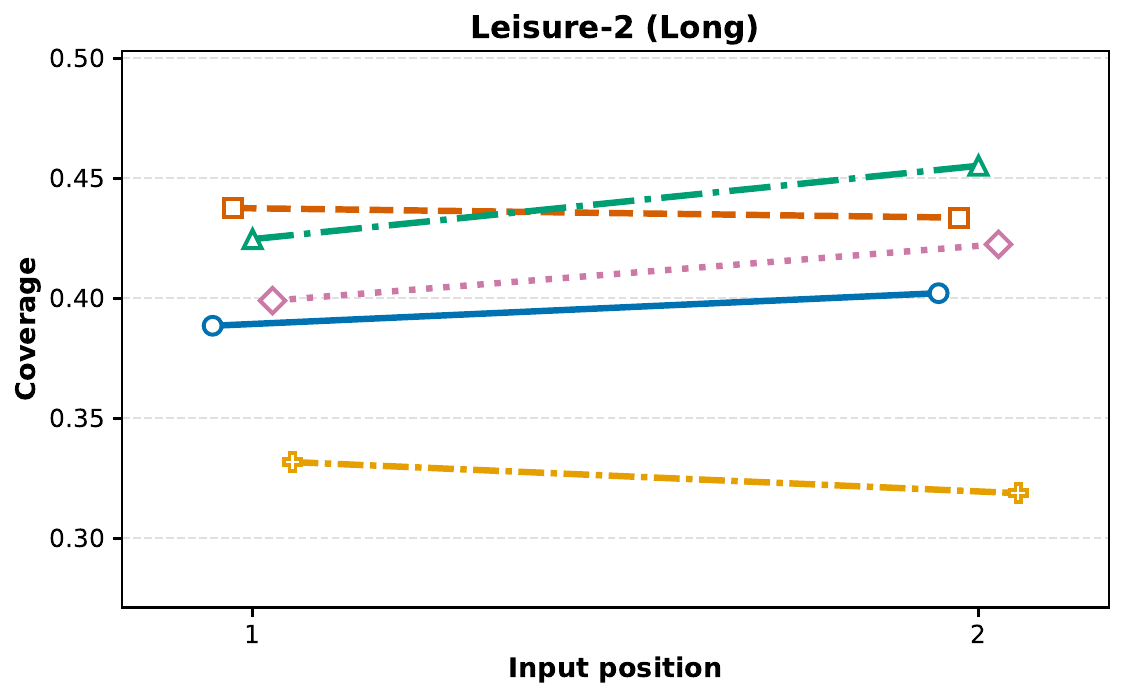}{Leisure-2 Long}

\vspace{1mm}

\curvepanel{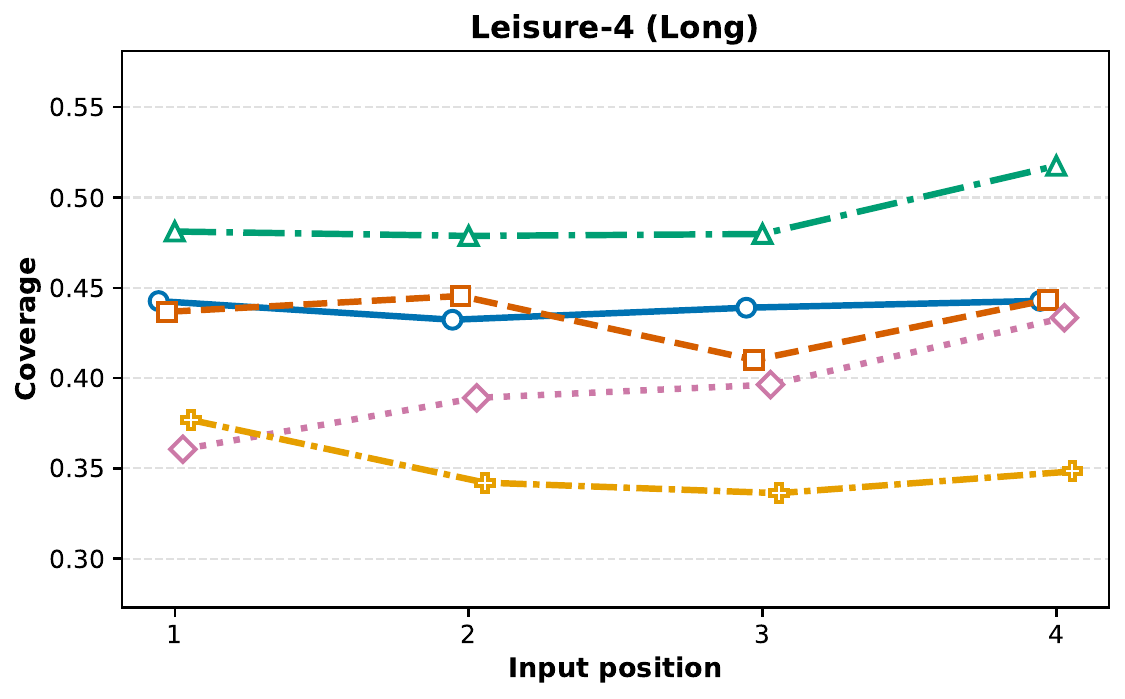}{Leisure-4 Long}
\hfill
\curvepanel{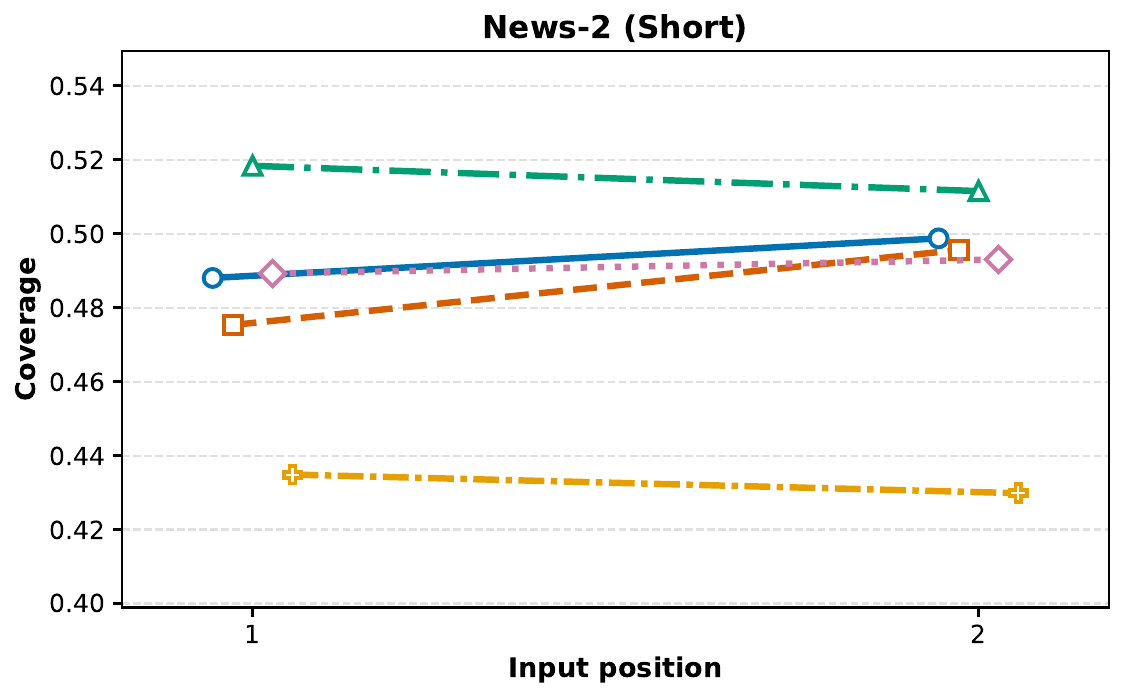}{News-2 Short}
\hfill
\curvepanel{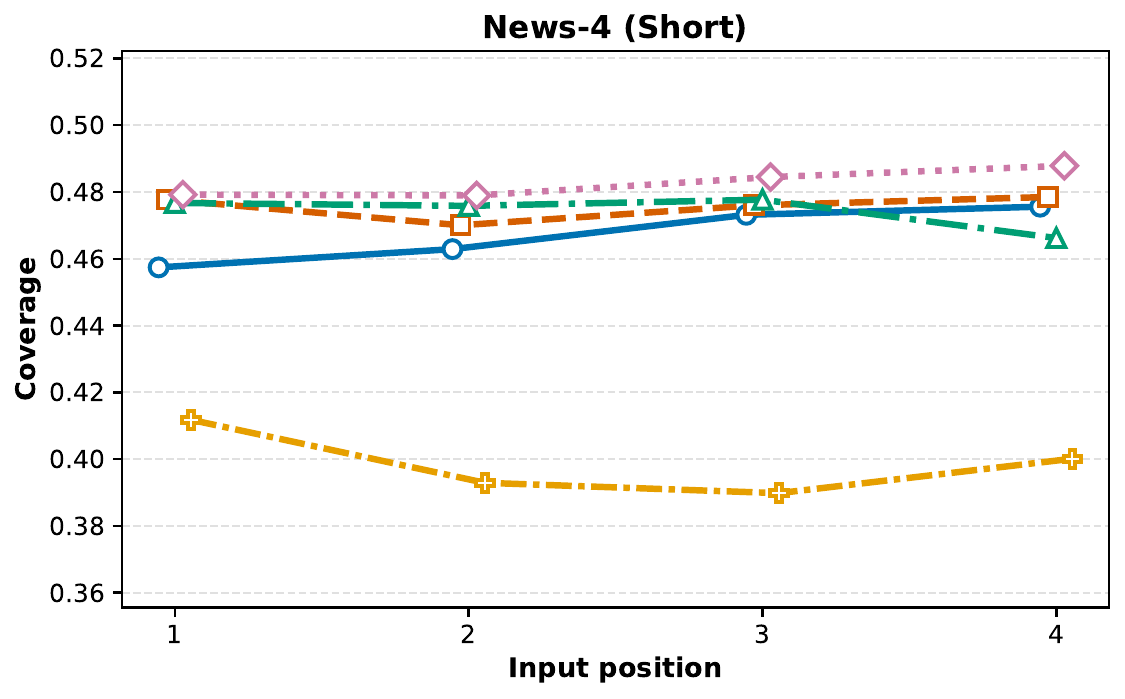}{News-4 Short}

\vspace{2mm}
{\scriptsize
\curvelegenditem{0072B2}{}{Qwen3-VL-8B}\quad
\curvelegenditem{D55E00}{}{Qwen3-VL-30B-A3B}\quad
\curvelegenditem{009E73}{}{GPT-5.4}\quad
\curvelegenditem{CC79A7}{}{Gemini-3.1-Pro}\quad
\curvelegenditem{E69F00}{}{GLM-4.1V-9B-Thinking}
}

\caption{
Baseline Coverage--position curves for model group A. Each panel corresponds
to one domain, duration, and input-size setting.}
\label{fig:base_position_curves_group_a}
\end{figure*}

\begin{figure*}[p]
\centering

\curvepanel{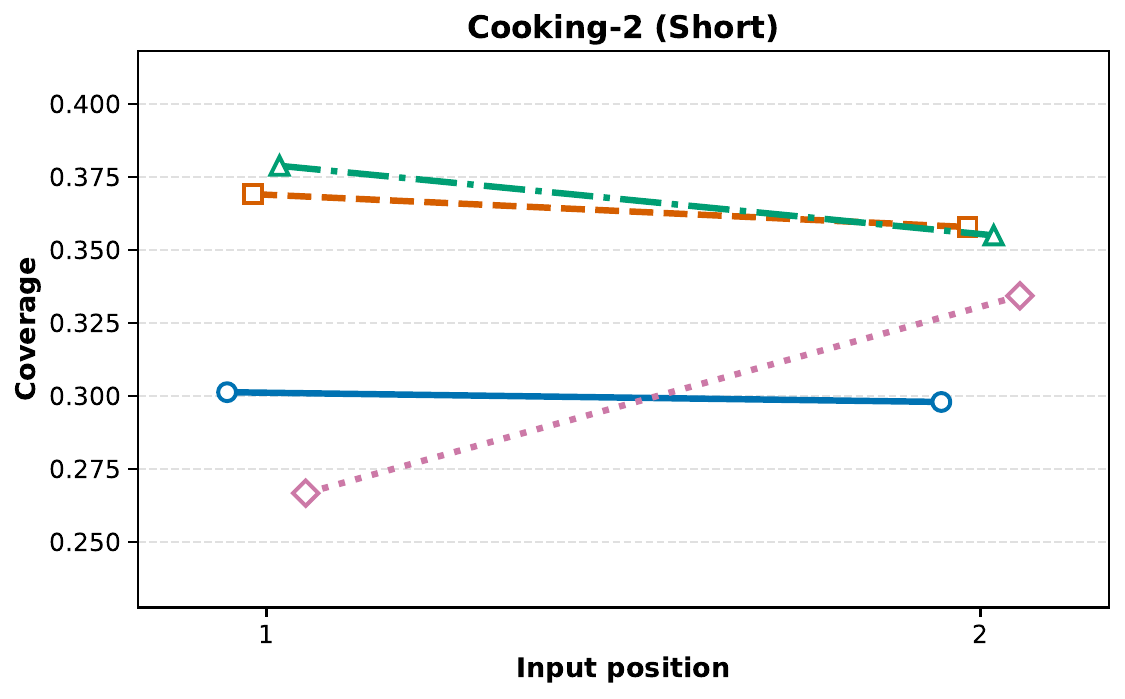}{Cooking-2 Short}
\hfill
\curvepanel{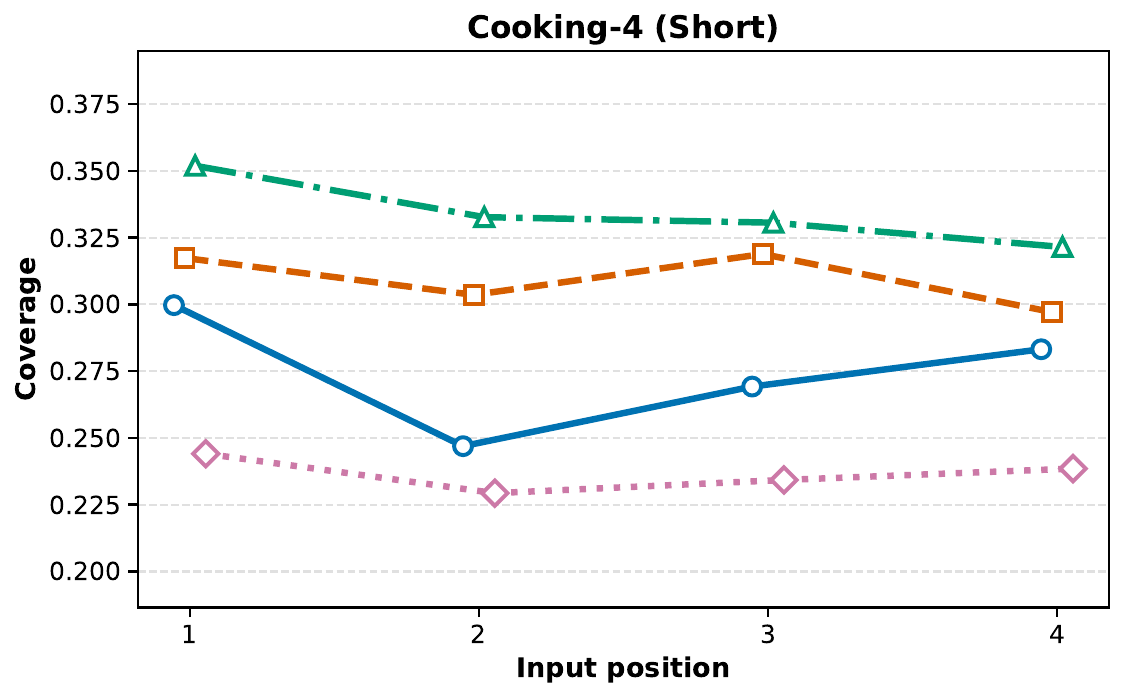}{Cooking-4 Short}
\hfill
\curvepanel{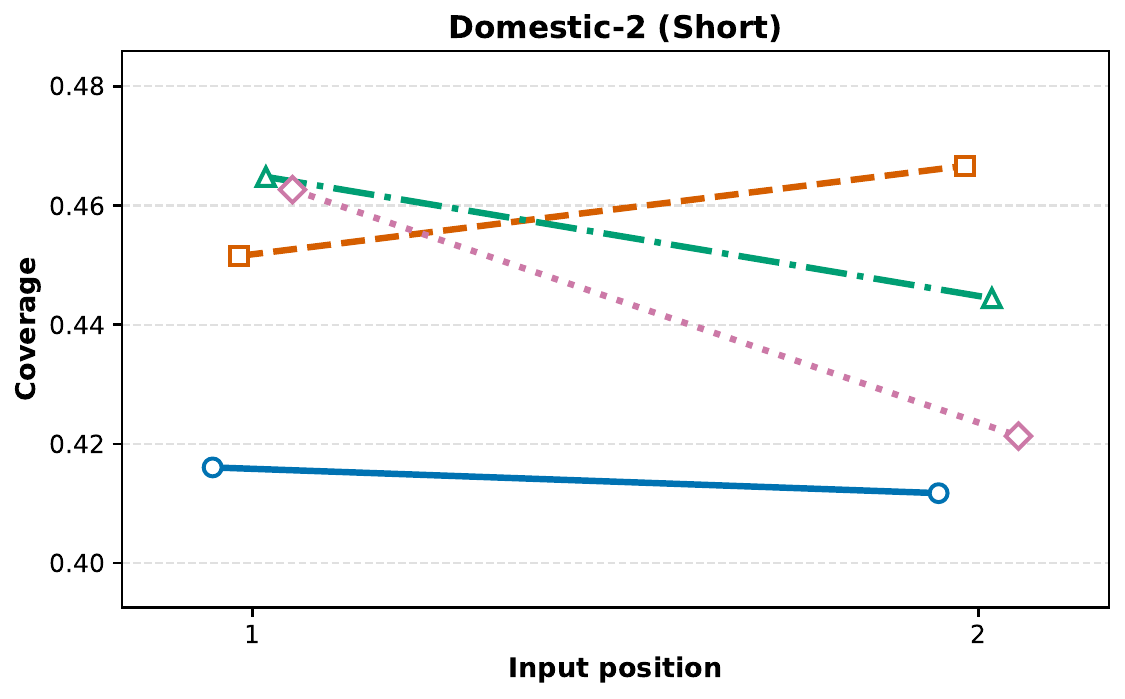}{Domestic-2 Short}

\vspace{1mm}

\curvepanel{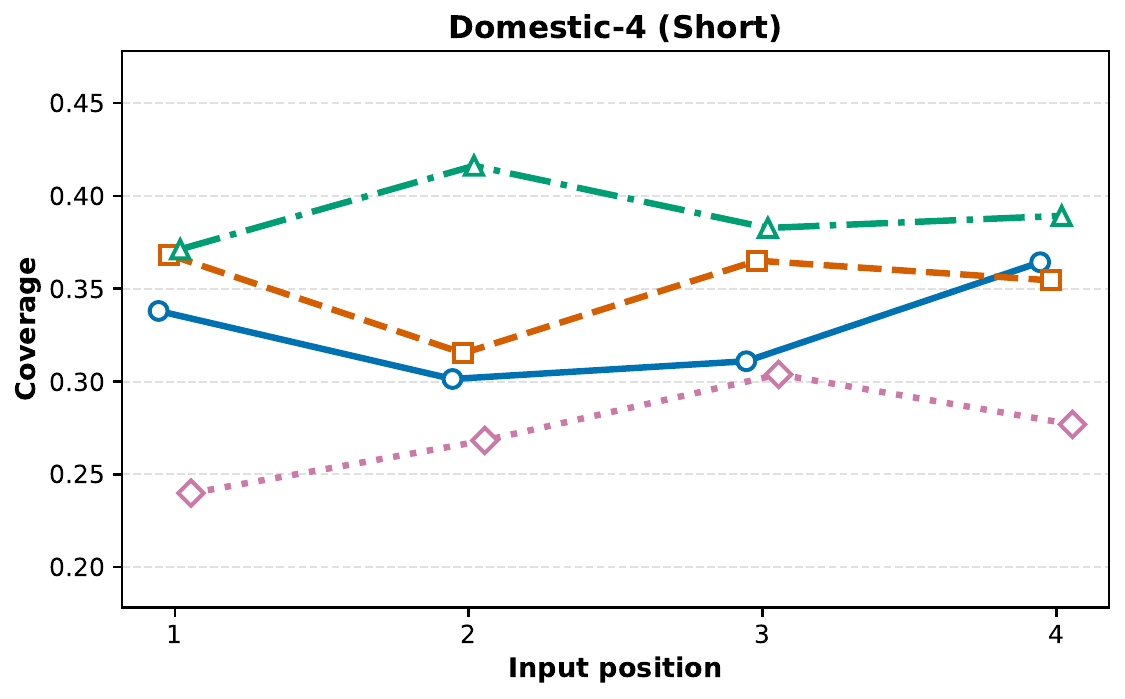}{Domestic-4 Short}
\hfill
\curvepanel{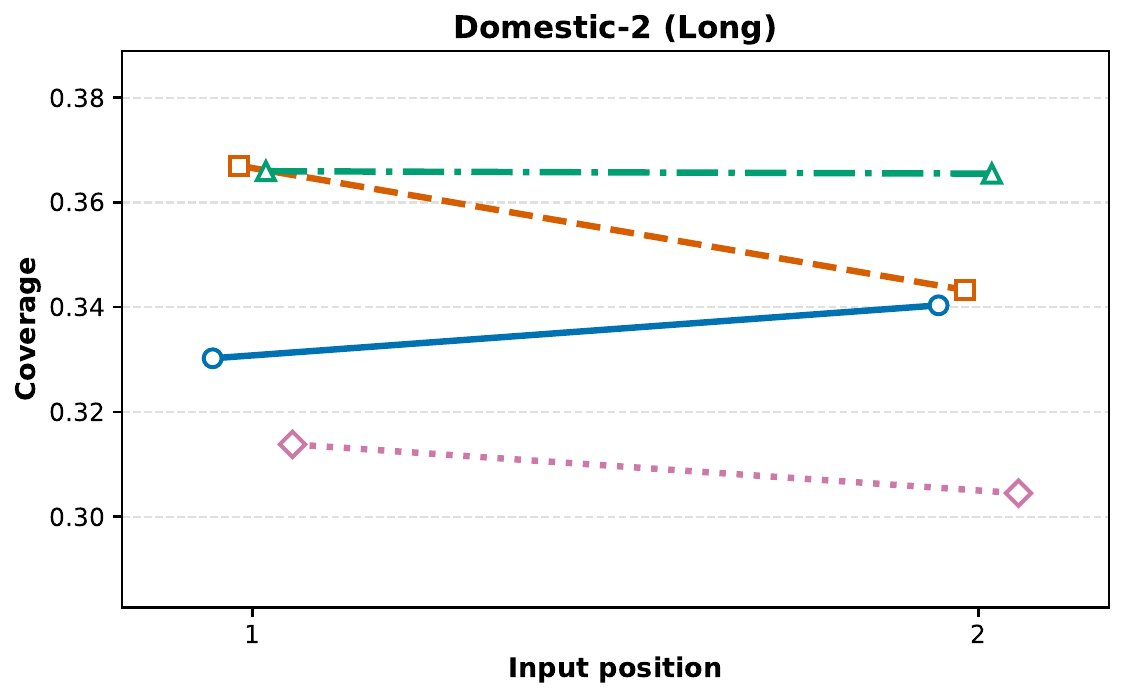}{Domestic-2 Long}
\hfill
\curvepanel{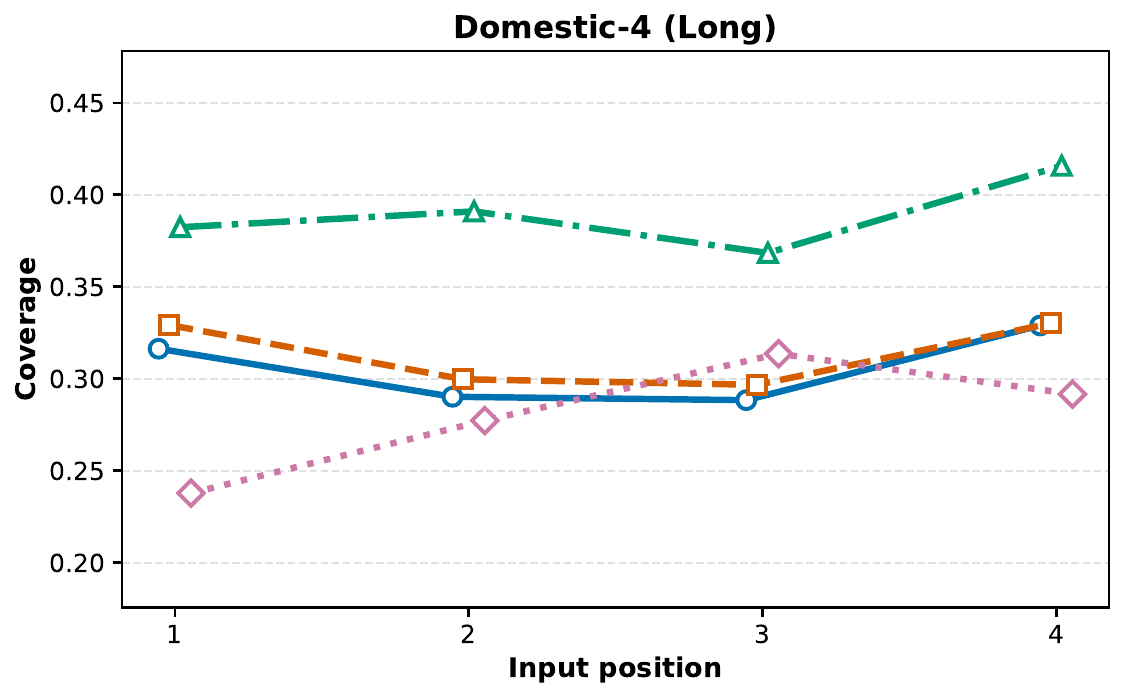}{Domestic-4 Long}

\vspace{1mm}

\curvepanel{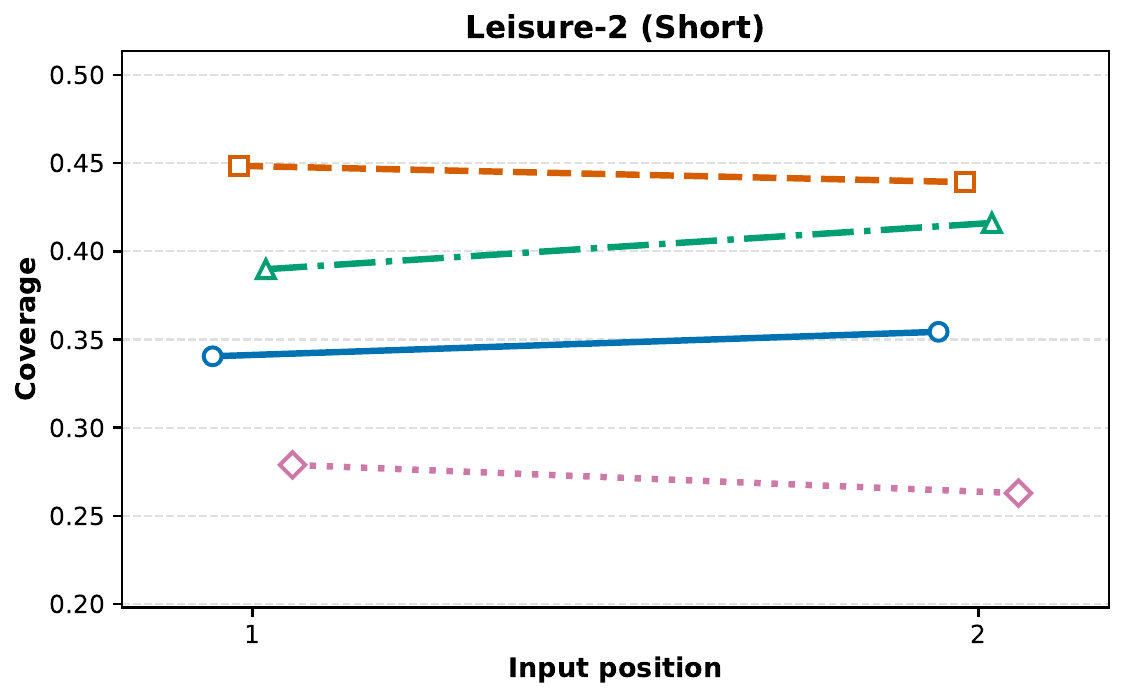}{Leisure-2 Short}
\hfill
\curvepanel{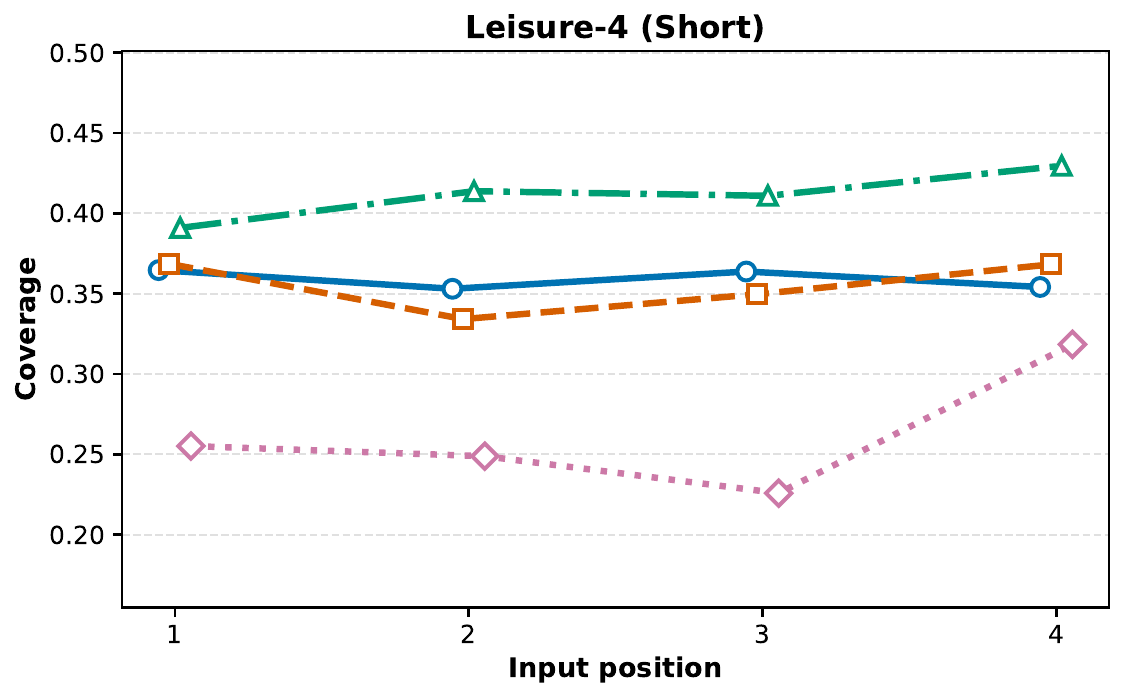}{Leisure-4 Short}
\hfill
\curvepanel{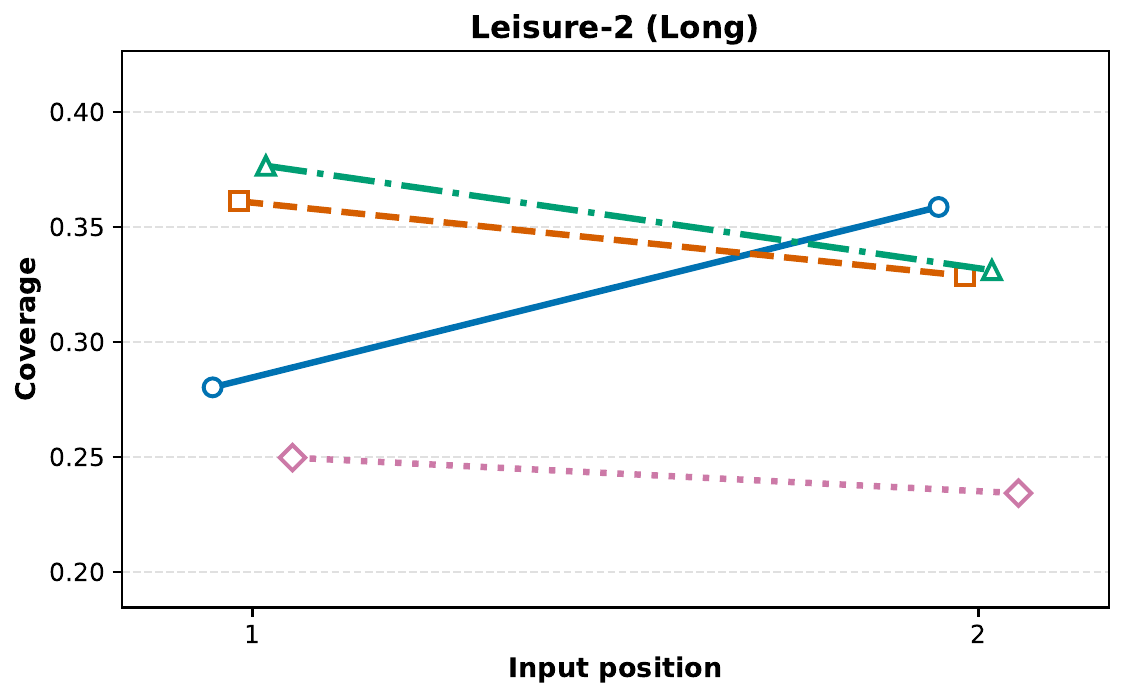}{Leisure-2 Long}

\vspace{1mm}

\curvepanel{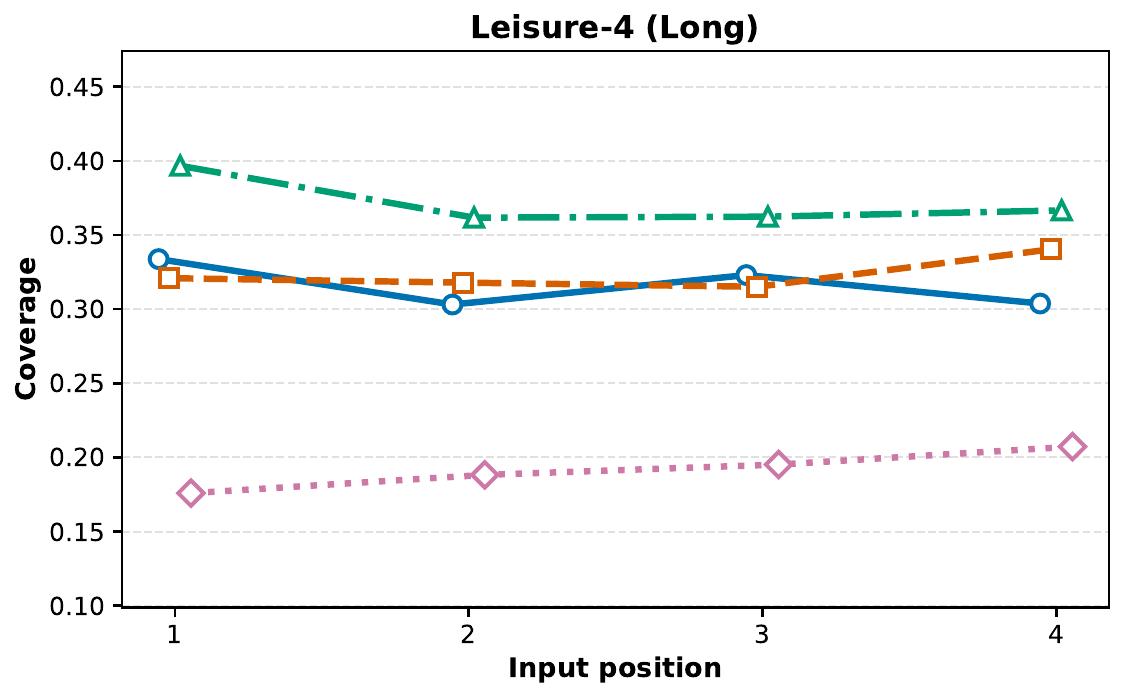}{Leisure-4 Long}
\hfill
\curvepanel{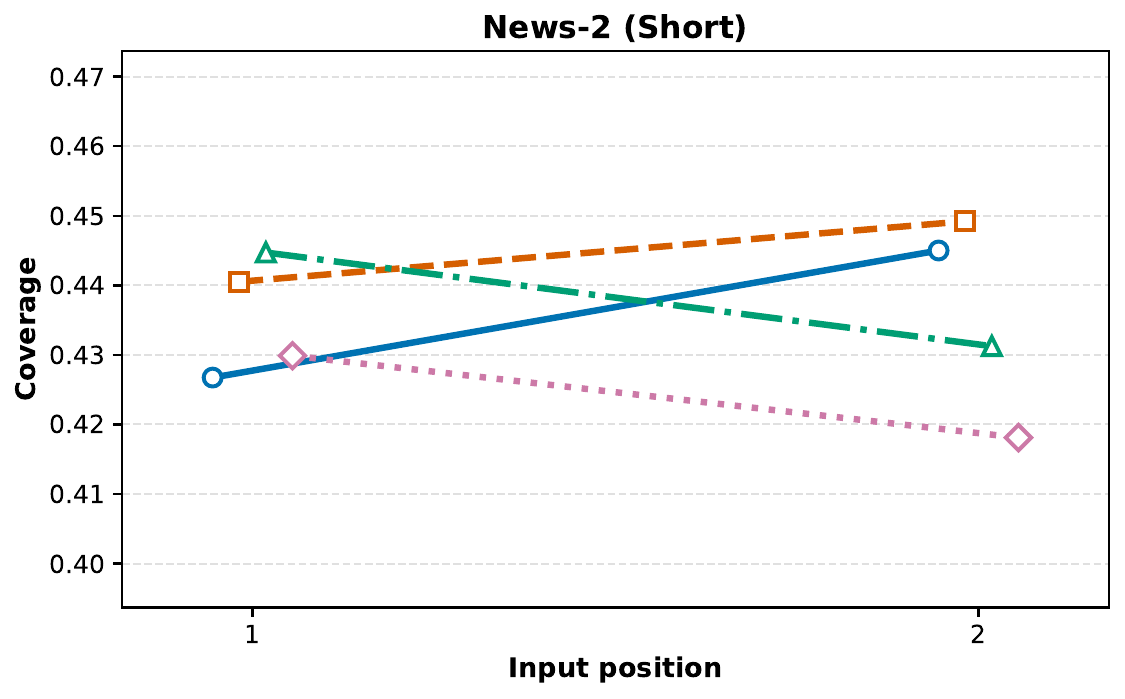}{News-2 Short}
\hfill
\curvepanel{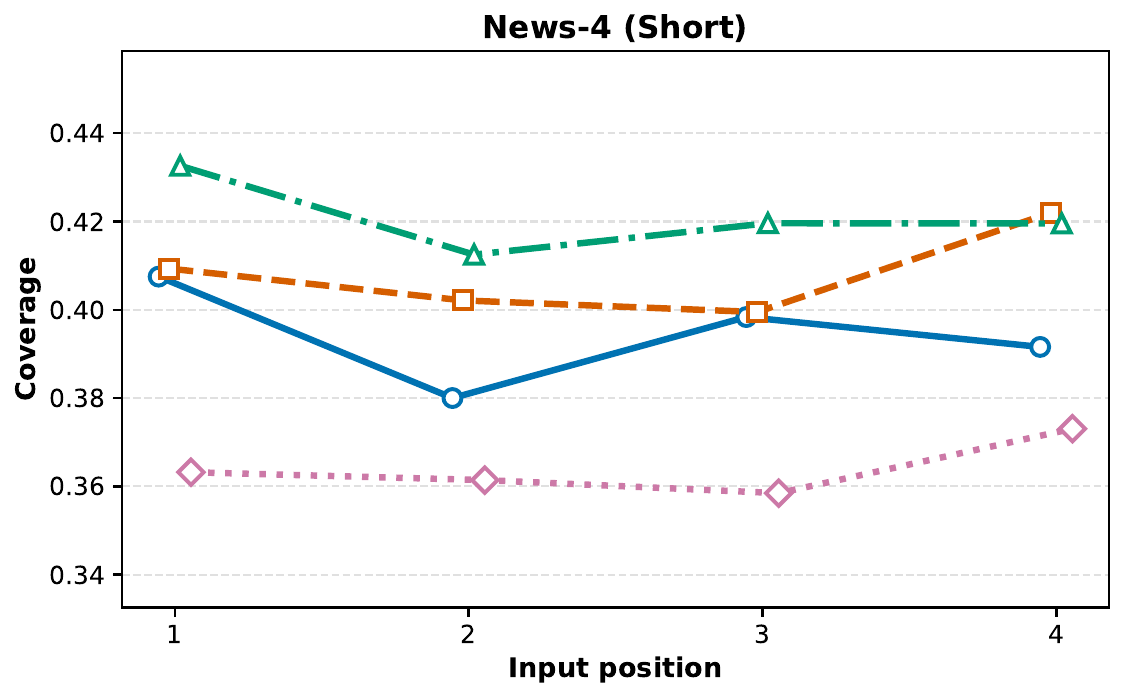}{News-4 Short}

\vspace{2mm}
{\scriptsize
\curvelegenditem{0072B2}{}{InternVL3.5-8B}\quad
\curvelegenditem{D55E00}{}{InternVL3.5-14B}\quad
\curvelegenditem{009E73}{}{MiniCPM-o-4.5}\quad
\curvelegenditem{CC79A7}{}{MiMo-VL-7B-RL}
}

\caption{
Baseline Coverage--position curves for model group B. The panel order matches
Figure~\ref{fig:base_position_curves_group_a}.}
\label{fig:base_position_curves_group_b}
\end{figure*}

\subsection{Visual-budget Ablation Tables}
\label{app:visual_budget_tables}

The main text reports a visual-budget robustness check for Qwen3-VL-8B. Here we
provide the full visual-budget summaries for Qwen3-VL-8B and MiniCPM-o-4.5,
including Coverage, DPB, and MEG. For Qwen3-VL-8B, negative MEG is largely
preserved across frame-count and resolution variants, especially in
\textit{Domestic}. MiniCPM-o-4.5 shows a different pattern: frame count mainly
changes overall Coverage, with 8 frames lowering Coverage and 24 frames
improving Coverage across domains, while MEG remains mixed. These results
support the main-text conclusion that larger visual budgets do not necessarily
remove positional imbalance.

\begin{table*}[t]
\centering
\scriptsize
\setlength{\tabcolsep}{2pt}
\renewcommand{\arraystretch}{0.92}

\begin{minipage}[t]{0.49\textwidth}
\centering
\textbf{(a) Qwen3-VL-8B}\\[1mm]
\resizebox{\linewidth}{!}{%
\begin{tabular}{llrrrrrrr}
\toprule
Domain & Metric & Base & 8F & 12F & 24F & 336R & 624R & 720R \\
\midrule
Cooking & Cov. & 39.48 & 32.07 & 36.73 & 38.18 & 35.62 & 41.03 & 38.90 \\
        & DPB  & -1.68 & 2.55 & -0.19 & -0.27 & 0.51 & 1.38 & 1.80 \\
        & MEG  & -1.34 & -3.59 & -3.17 & -2.06 & -3.45 & 0.63 & -3.83 \\
\midrule
Domestic & Cov. & 42.11 & 37.32 & 40.47 & 42.59 & 39.01 & 41.09 & 42.64 \\
         & DPB  & -0.72 & -2.70 & 0.69 & 0.07 & -1.17 & 3.57 & -0.81 \\
         & MEG  & -2.57 & -2.20 & -3.56 & -3.87 & -0.56 & -0.70 & -5.50 \\
\midrule
Leisure & Cov. & 43.79 & 38.08 & 39.81 & 43.68 & 42.33 & 44.23 & 43.71 \\
        & DPB  & -0.18 & 2.22 & -2.23 & 0.37 & 0.48 & 0.83 & 0.53 \\
        & MEG  & -0.64 & -1.13 & -1.49 & -0.19 & -3.12 & -1.89 & -0.10 \\
\midrule
News & Cov. & 46.73 & 37.14 & 44.78 & 47.04 & 45.92 & 46.27 & 45.87 \\
     & DPB  & 1.16 & 0.07 & 0.95 & 0.85 & 0.51 & 0.21 & 0.37 \\
     & MEG  & 0.15 & 1.27 & 0.49 & -0.24 & 0.34 & -1.00 & 1.30 \\
\bottomrule
\end{tabular}}
\end{minipage}
\hfill
\begin{minipage}[t]{0.49\textwidth}
\centering
\textbf{(b) MiniCPM-o-4.5}\\[1mm]
\resizebox{\linewidth}{!}{%
\begin{tabular}{llrrrrrrr}
\toprule
Domain & Metric & Base & 8F & 12F & 24F & 336R & 624R & 720R \\
\midrule
Cooking & Cov. & 34.23 & 30.01 & 34.48 & 38.04 & 32.70 & 34.93 & 35.49 \\
        & DPB  & -0.58 & 0.49 & 1.00 & -0.39 & -0.30 & 0.39 & -0.42 \\
        & MEG  & -2.03 & -2.42 & -1.79 & 0.40 & -0.65 & -3.06 & 1.02 \\
\midrule
Domestic & Cov. & 39.35 & 34.38 & 34.96 & 40.85 & 41.93 & 41.84 & 40.83 \\
         & DPB  & 0.79 & 0.90 & -2.27 & 0.47 & -0.35 & -3.42 & 1.61 \\
         & MEG  & 0.86 & -1.45 & 0.52 & -1.35 & -0.58 & -1.48 & 3.31 \\
\midrule
Leisure & Cov. & 40.90 & 35.00 & 39.54 & 42.70 & 40.42 & 40.89 & 40.29 \\
        & DPB  & -0.32 & -2.13 & -1.84 & -1.10 & 0.64 & -0.43 & -0.00 \\
        & MEG  & -0.85 & 0.81 & -0.96 & -1.05 & -1.34 & -2.36 & 0.01 \\
\midrule
News & Cov. & 42.05 & 31.20 & 39.01 & 42.51 & 41.54 & 41.43 & 42.05 \\
     & DPB  & -0.45 & -2.45 & -1.17 & 0.19 & -0.60 & -0.00 & -0.58 \\
     & MEG  & -1.19 & -0.78 & 0.60 & -1.18 & -0.16 & -1.04 & -0.10 \\
\bottomrule
\end{tabular}}
\end{minipage}

\caption{
Visual-budget ablation summary for Qwen3-VL-8B and MiniCPM-o-4.5 on short
$P{=}4$ settings. All values are percentages. Base denotes 16 frames at
$448{\times}448$. Frame-count variants fix resolution at $448{\times}448$;
resolution variants fix frame count at 16.
}
\label{tab:visual_budget_full_table}
\end{table*}
\subsection{InternVL3.5-14B Single-target All-dataset Results}
\label{app:intern14_single_target_all_datasets}

Tables~\ref{tab:intern14_single_target_mean_coverage_gain} and
\ref{tab:intern14_single_target_weak_position_gain} provide the dataset-level
results behind the single-target analysis in
Section~\ref{sec:single_target_prompting}. The tables separate mean Coverage
changes from weak-position changes. Single-target prompting lowers mean Coverage
in all six $P{=}2$ settings, but improves mean Coverage in all six $P{=}4$
settings. The weak-position analysis shows the same scale dependence: none of
the $P{=}2$ weak positions improves, while 4 of 6 $P{=}4$ weak positions improve.

\begin{table}[t]
\centering
\small
\setlength{\tabcolsep}{4pt}
\begin{tabular}{llcrrr}
\toprule
Domain & Dur. & $P$ & Joint & Single & Gain \\
\midrule
Cooking & S & 2 & 0.3635 & 0.3113 & $-5.22$ \\
Cooking & S & 4 & 0.3092 & 0.3181 & $+0.88$ \\
Domestic & L & 2 & 0.3551 & 0.3193 & $-3.59$ \\
Domestic & L & 4 & 0.3139 & 0.3148 & $+0.09$ \\
Domestic & S & 2 & 0.4591 & 0.4289 & $-3.02$ \\
Domestic & S & 4 & 0.3508 & 0.3617 & $+1.09$ \\
News & S & 2 & 0.4449 & 0.4384 & $-0.65$ \\
News & S & 4 & 0.4082 & 0.4350 & $+2.68$ \\
Leisure & L & 2 & 0.3449 & 0.3074 & $-3.75$ \\
Leisure & L & 4 & 0.3235 & 0.3287 & $+0.52$ \\
Leisure & S & 2 & 0.4439 & 0.3875 & $-5.63$ \\
Leisure & S & 4 & 0.3552 & 0.4006 & $+4.54$ \\
\bottomrule
\end{tabular}
\caption{
Dataset-level mean Coverage for InternVL3.5-14B under joint generation and
single-target prompting. Gain is single-target minus joint generation, reported
in percentage points.
}
\label{tab:intern14_single_target_mean_coverage_gain}
\end{table}

\begin{table}[t]
\centering
\small
\setlength{\tabcolsep}{4pt}
\resizebox{\linewidth}{!}{%
\begin{tabular}{llcrrr}
\toprule
Domain & Dur. & Weak pos. & Weak gain & Nonweak gain & Diff. \\
\midrule
Cooking & S & pos4 & $+5.29$ & $-0.59$ & $+5.88$ \\
Domestic & L & pos3 & $-0.07$ & $+0.15$ & $-0.22$ \\
Domestic & S & pos2 & $+3.66$ & $+0.23$ & $+3.43$ \\
News & S & pos3 & $+4.15$ & $+2.18$ & $+1.97$ \\
Leisure & L & pos3 & $-0.90$ & $+0.99$ & $-1.89$ \\
Leisure & S & pos2 & $+5.07$ & $+4.36$ & $+0.71$ \\
\bottomrule
\end{tabular}
}
\caption{
Weak-position gain for InternVL3.5-14B in $P{=}4$ settings. The weak position is
the lowest-Coverage position under joint generation. Gains are single-target
minus joint generation, reported in percentage points. Diff. is weak-position
gain minus the average gain of the remaining positions.
}
\label{tab:intern14_single_target_weak_position_gain}
\end{table}

\subsection{Equal-attention Coverage Curves}
\label{app:equal_attention_curves}

\begin{figure*}[t]
\centering
\includegraphics[width=\textwidth]{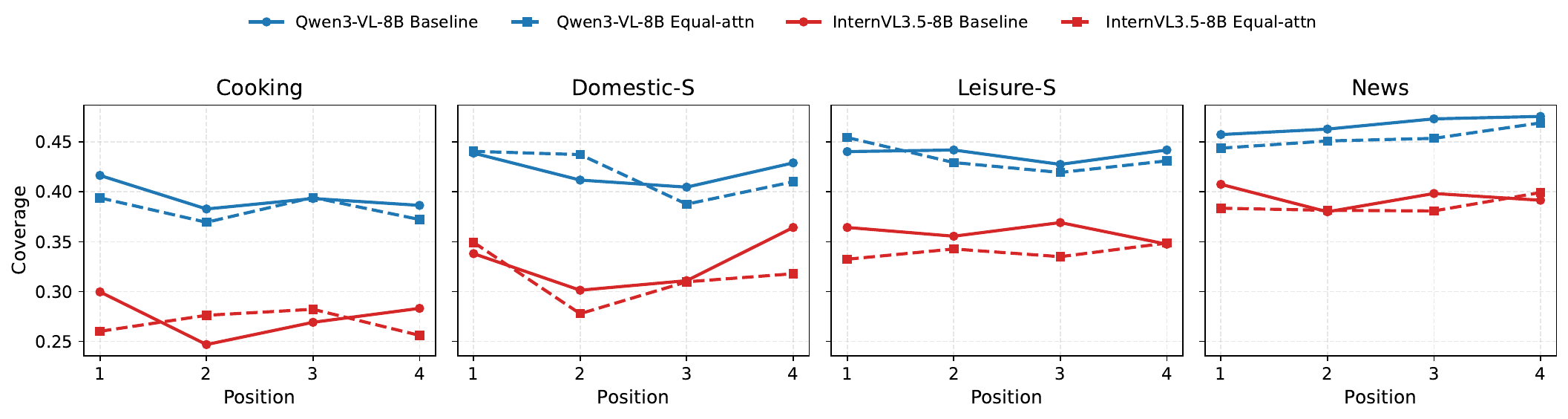}
\caption{\textbf{Equal-attention prompt Coverage curves across all short four-video domains.}
Solid lines are baseline prompt-first black-frame runs; dashed lines add the explicit equal-attention instruction.}
\label{fig:equal_attention_all_curves}
\end{figure*}

Figure~\ref{fig:equal_attention_all_curves} reports the equal-attention prompt
intervention on all four short four-video domains for Qwen3-VL-8B and
InternVL3.5-8B. These curves complement the main-text two-domain visualization
and show that the prompt intervention is model- and domain-dependent rather than
a uniform fix for positional imbalance.

\subsection{Cyclic Versus Full-Permutation Ordering}
\label{app:permutation_curves}

The main experiments use cyclic orderings rather than full permutations. This design ensures that each video appears at each absolute position exactly once, while keeping inference and evaluation cost manageable. This trade-off is important: the current study already required roughly 1,100 hours for model inference and evaluation, and a full-permutation design would multiply the four-video setting from 4 cyclic orders to \(4! = 24\) orders before considering models, domains, robustness settings, and judge-based scoring. The limitation is that cyclic ordering cannot fully average over surrounding context: absolute position effects may still be partially entangled with which videos appear before or after the target video.

\begin{figure*}[t]
\centering
\includegraphics[width=0.78\textwidth]{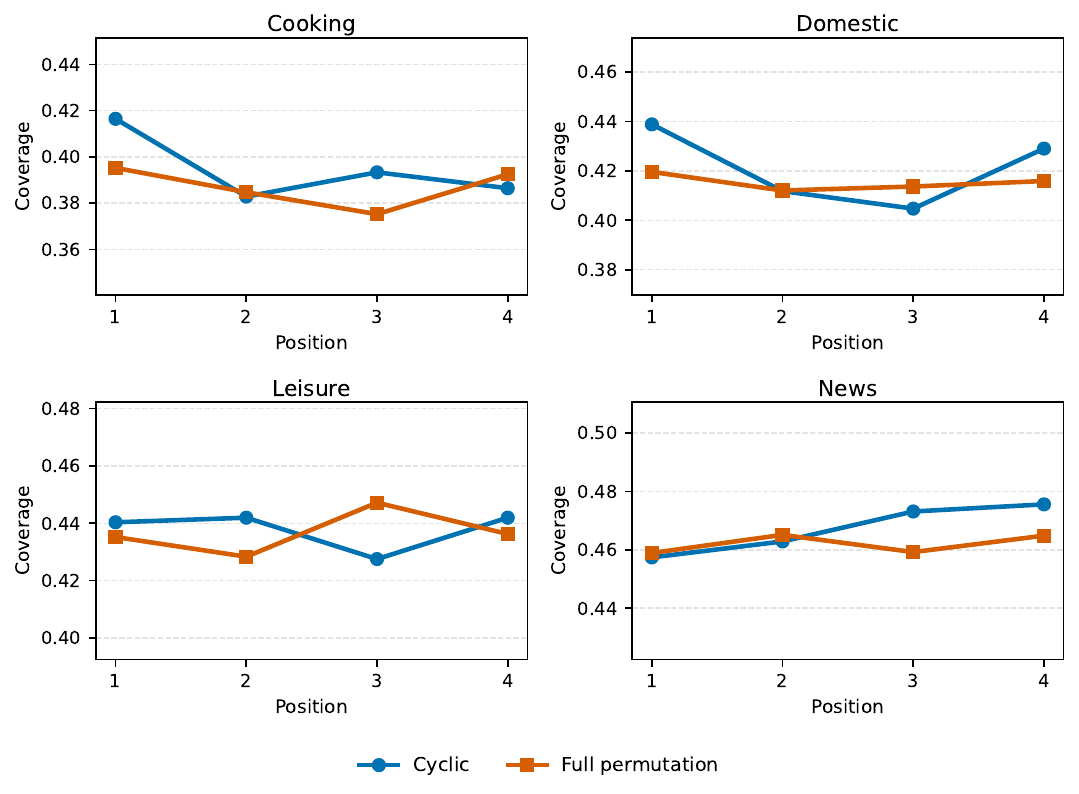}
\caption{\textbf{Qwen3-VL-8B cyclic versus full-permutation Coverage curves.}
The full-permutation design averages over all order contexts for each target position.}
\label{fig:qwen8_cyclic_full_permutation}
\end{figure*}

\begin{table}[H]
\centering
\small
\setlength{\tabcolsep}{5pt}
\begin{tabular}{lccc}
\toprule
Dataset & Cyclic MEG & Full MEG & Mean $|\Delta_{\mathrm{pos}}|$ \\
\midrule
Cooking & $-1.34$ & $-1.39$ & $1.18$ \\
Domestic & $-2.57$ & $-0.49$ & $1.04$ \\
Leisure & $-0.64$ & $+0.20$ & $1.10$ \\
News & $+0.15$ & $+0.03$ & $0.71$ \\
\bottomrule
\end{tabular}
\caption{\textbf{Cyclic versus full-permutation comparison for Qwen3-VL-8B.}
MEG and mean absolute position-wise Coverage differences are reported in percentage points.}
\label{tab:qwen8_cyclic_full_summary}
\end{table}

To assess this trade-off, Figure~\ref{fig:qwen8_cyclic_full_permutation} compares Qwen3-VL-8B under the cyclic design and a full-permutation design on the short four-video setting. Full permutation separates three sources that are partially entangled in cyclic runs: absolute target position, surrounding video-order context, and residual generation or evaluation noise. The full-permutation results make the cyclic middle-position weakness slightly weaker on average: mean MEG changes from \(-1.10\) percentage points under cyclic ordering to \(-0.41\) percentage points under full permutation. However, the cyclic estimate remains a useful approximation for the main positional trend. The average absolute difference between cyclic and full-permutation position means is small, ranging from \(0.71\) to \(1.18\) percentage points across domains. The strongest cyclic middle weakness appears in Domestic, where full permutation reduces the magnitude but does not turn it into a strong middle advantage. Cooking remains negative under both designs, while Leisure and News are near zero under both. Overall, the check suggests that cyclic ordering can slightly overstate middle weakness in some domains, but does not produce a qualitatively different position curve in the current Qwen3-VL-8B setting.

\section{Examples}

\begin{figure*}[t]
\centering

\begin{tcolorbox}[
    width=0.98\textwidth,
    colback=blue!2,
    colframe=blue!35,
    boxrule=0.8pt,
    arc=2mm,
    left=3mm,
    right=3mm,
    top=2mm,
    bottom=2mm
]

\begin{tcolorbox}[
    colback=blue!15,
    colframe=blue!15,
    boxrule=0pt,
    arc=2mm,
    left=2mm,
    right=2mm,
    top=1mm,
    bottom=1mm
]
{\Large \textbf{Example 1.}}
\end{tcolorbox}

\vspace{2mm}

\begin{center}
\fcolorbox{blue!35}{white}{
    \includegraphics[width=0.88\linewidth]{figs/Example1.pdf}
}
\end{center}

\vspace{2mm}

\noindent
\begin{minipage}[t]{0.232\linewidth}
\begin{tcolorbox}[
    colback=white,
    colframe=blue!45,
    title=\textbf{Position 1},
    coltitle=black,
    colbacktitle=blue!15,
    fonttitle=\bfseries,
    boxrule=0.7pt,
    arc=1.5mm,
    left=1.2mm,
    right=1.2mm,
    top=1mm,
    bottom=1mm,
    height=7.8cm
]
\small
\RaggedRight
A man demonstrates how to start a fire in a campground fire pit using a
\goodmark{\checkmark\ fire starter}.
He strikes the rod several times over a pile of tinder, which includes wood shavings and lichen, creating sparks.
After a few tries, the tinder
\goodmark{\checkmark\ catches fire},
and he successfully starts a small flame while his
\goodmark{\checkmark\ children watch}.
\end{tcolorbox}
\end{minipage}\hfill
\begin{minipage}[t]{0.232\linewidth}
\begin{tcolorbox}[
    colback=white,
    colframe=blue!45,
    title=\textbf{Position 2},
    coltitle=black,
    colbacktitle=blue!15,
    fonttitle=\bfseries,
    boxrule=0.7pt,
    arc=1.5mm,
    left=1.2mm,
    right=1.2mm,
    top=1mm,
    bottom=1mm,
    height=7.8cm
]
\small
\RaggedRight
A man at a campsite is preparing to start a fire inside a metal fire ring.
He uses a
\goodmark{\checkmark\ fire striker}
to create sparks on a tinder bundle made of wood shavings and some white material.
After a few tries, the tinder
\goodmark{\checkmark\ catches fire},
and he successfully starts a small flame.
\end{tcolorbox}
\end{minipage}\hfill
\begin{minipage}[t]{0.232\linewidth}
\begin{tcolorbox}[
    colback=white,
    colframe=blue!45,
    title=\textbf{Position 3},
    coltitle=black,
    colbacktitle=blue!15,
    fonttitle=\bfseries,
    boxrule=0.7pt,
    arc=1.5mm,
    left=1.2mm,
    right=1.2mm,
    top=1mm,
    bottom=1mm,
    height=7.8cm
]
\small
\RaggedRight
A man demonstrates how to start a campfire using a
\goodmark{\checkmark\ fire starter}
in a metal fire pit.
He scrapes a ferro rod over a tinder bundle of wood shavings and lichen, creating sparks to ignite it.
After a few strikes,
\goodmark{\checkmark\ the sparks catch},
and he successfully starts a small flame.
\end{tcolorbox}
\end{minipage}\hfill
\begin{minipage}[t]{0.232\linewidth}
\begin{tcolorbox}[
    colback=white,
    colframe=blue!45,
    title=\textbf{Position 4},
    coltitle=black,
    colbacktitle=blue!15,
    fonttitle=\bfseries,
    boxrule=0.7pt,
    arc=1.5mm,
    left=1.2mm,
    right=1.2mm,
    top=1mm,
    bottom=1mm,
    height=7.8cm
]
\small
\RaggedRight
A man at a campsite demonstrates how to start a fire using a
\goodmark{\checkmark\ ferro rod}
and tinder.
He carefully arranges wood shavings and other flammable materials in a fire pit before striking the rod to create sparks.
After a few tries, the sparks successfully ignite the tinder, and
\goodmark{\checkmark\ a small flame}
begins to grow.
\end{tcolorbox}
\end{minipage}

\end{tcolorbox}

\caption{\textbf{Example 1.}
The four responses consistently preserve the core ignition-related details (e.g., \textit{fire starter} and \textit{catches fire}), while the contextual detail about children watching appears only in Position~1.}
\label{fig:example_fire_positions_a}
\end{figure*}

\begin{figure*}[t]
\centering
\begin{tcolorbox}[
    width=0.98\textwidth,
    colback=blue!2,
    colframe=blue!35,
    boxrule=0.8pt,
    arc=2mm,
    left=3mm,
    right=3mm,
    top=2mm,
    bottom=2mm
]

\begin{tcolorbox}[
    colback=blue!15,
    colframe=blue!15,
    boxrule=0pt,
    arc=2mm,
    left=2mm,
    right=2mm,
    top=1mm,
    bottom=1mm
]
{\large \textbf{Prompt-placement leakage example.}}
\end{tcolorbox}

\vspace{1.5mm}

\noindent\textbf{Claimed target slot:} Position 4, ``Drinking coffee''
\hfill

\vspace{1mm}
\noindent
\includegraphics[width=0.238\linewidth]{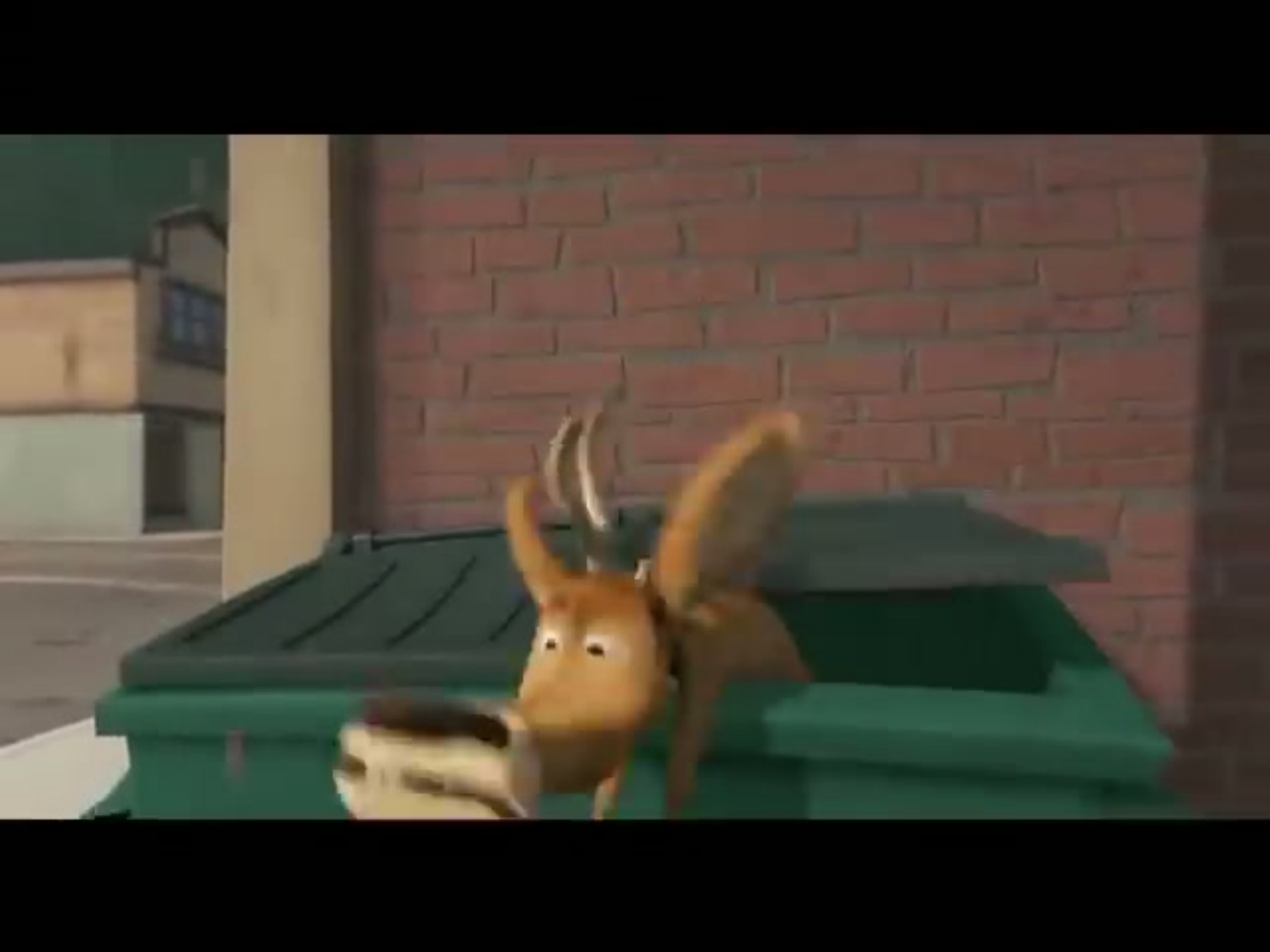}\hfill
\includegraphics[width=0.238\linewidth]{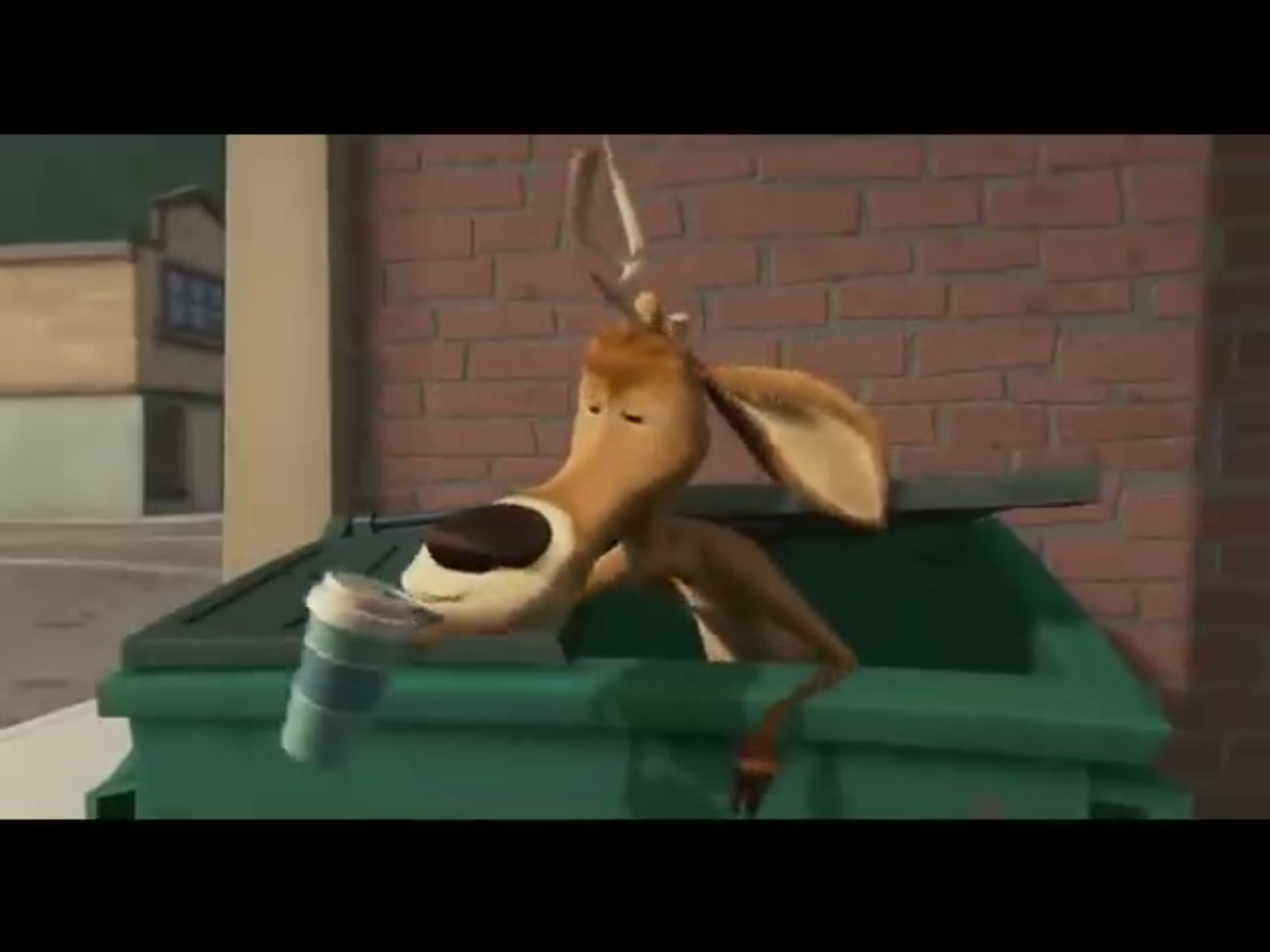}\hfill
\includegraphics[width=0.238\linewidth]{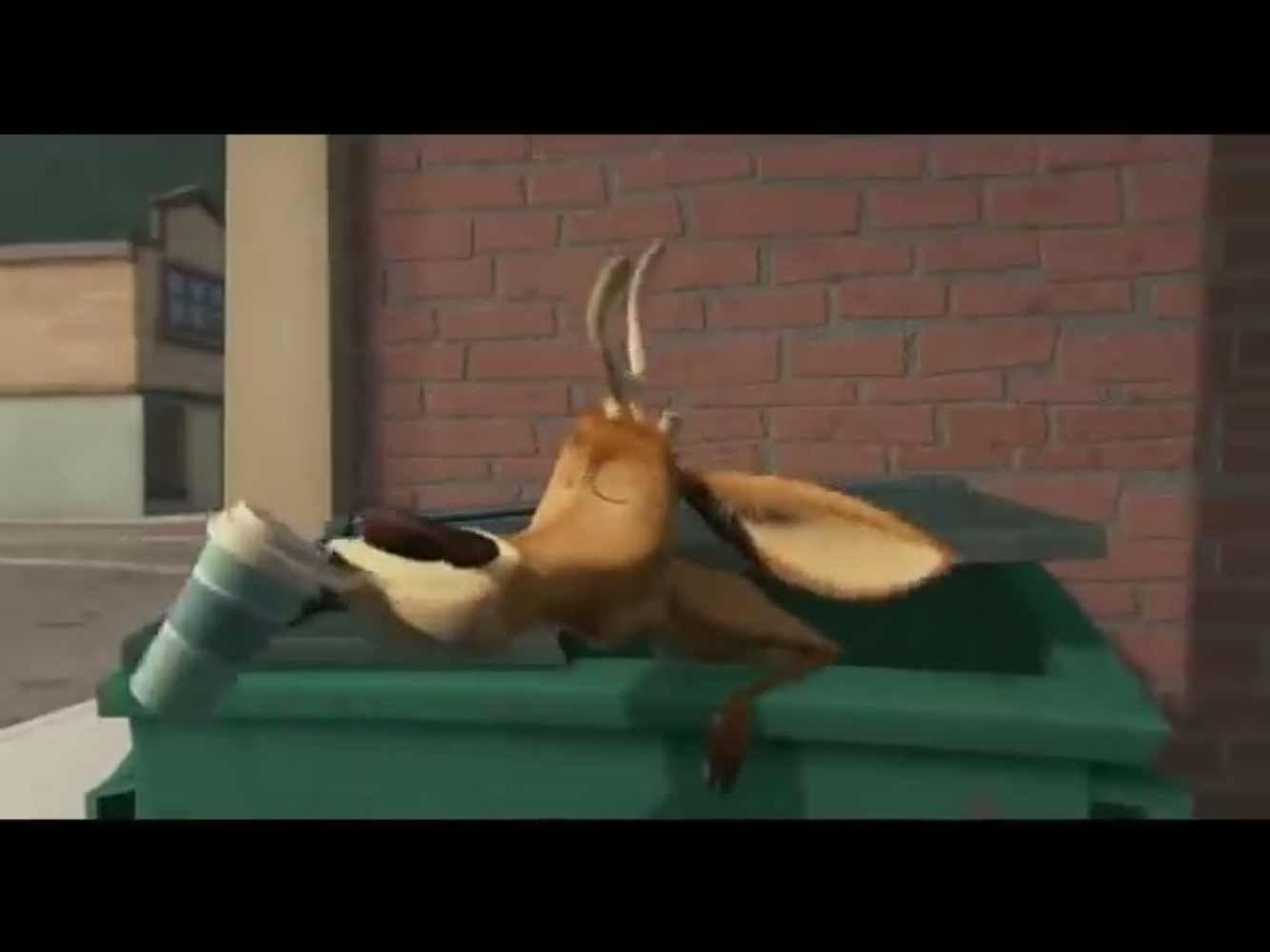}\hfill
\includegraphics[width=0.238\linewidth]{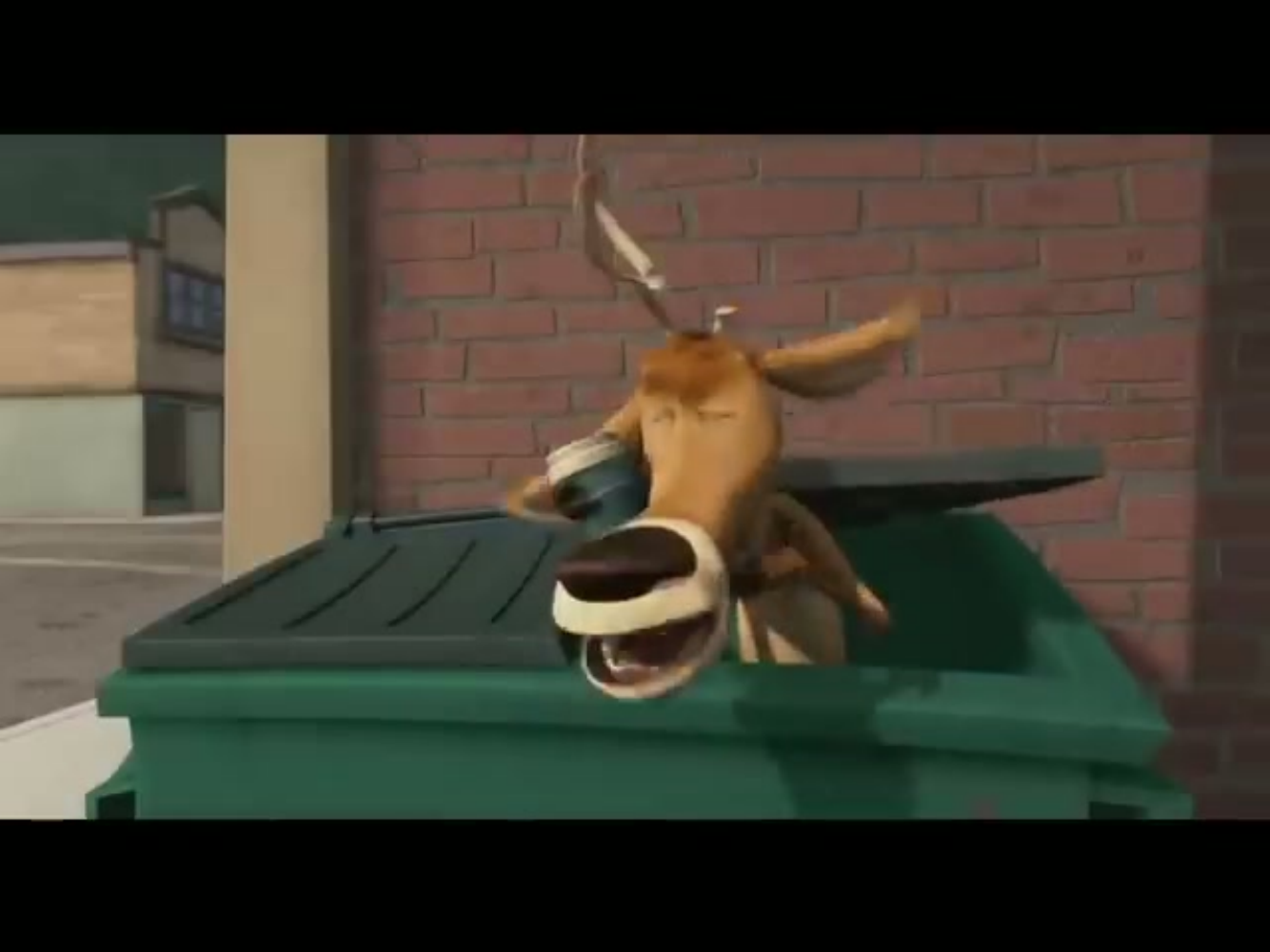}

\vspace{2mm}

\noindent\textbf{Source slot whose details leak in:} Position 1, ``Having an ice cream''
\hfill

\vspace{1mm}
\noindent
\includegraphics[width=0.238\linewidth]{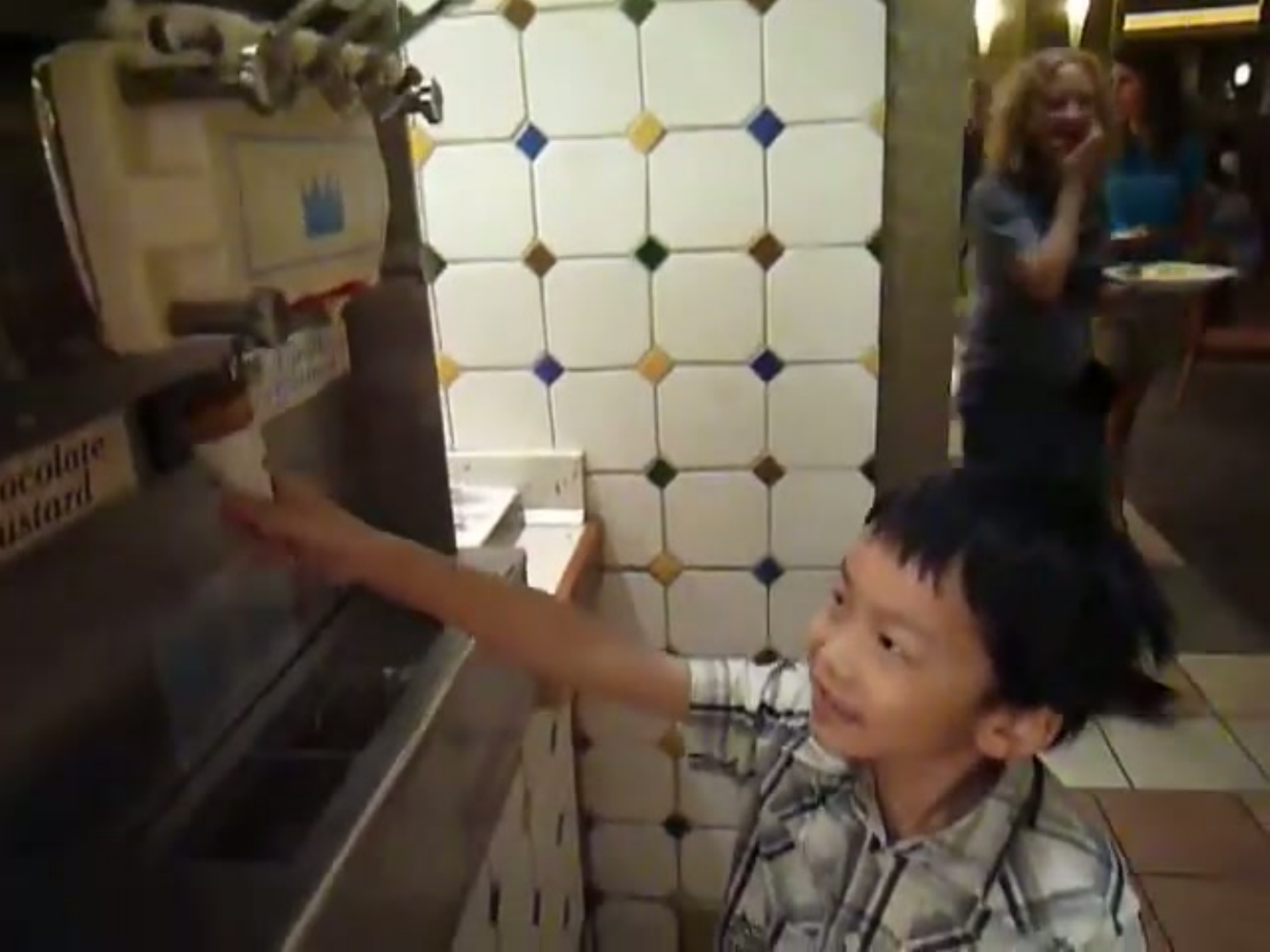}\hfill
\includegraphics[width=0.238\linewidth]{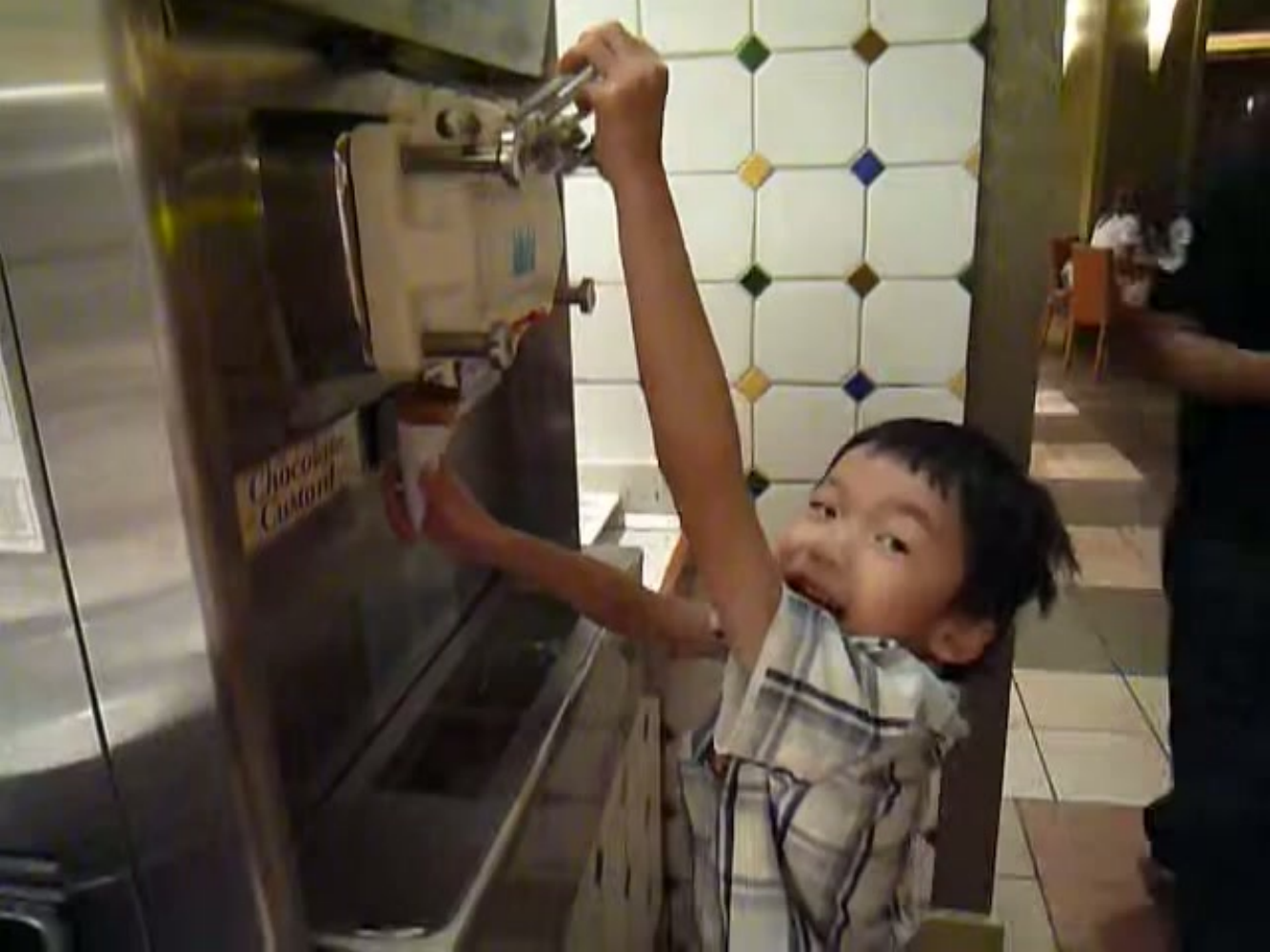}\hfill
\includegraphics[width=0.238\linewidth]{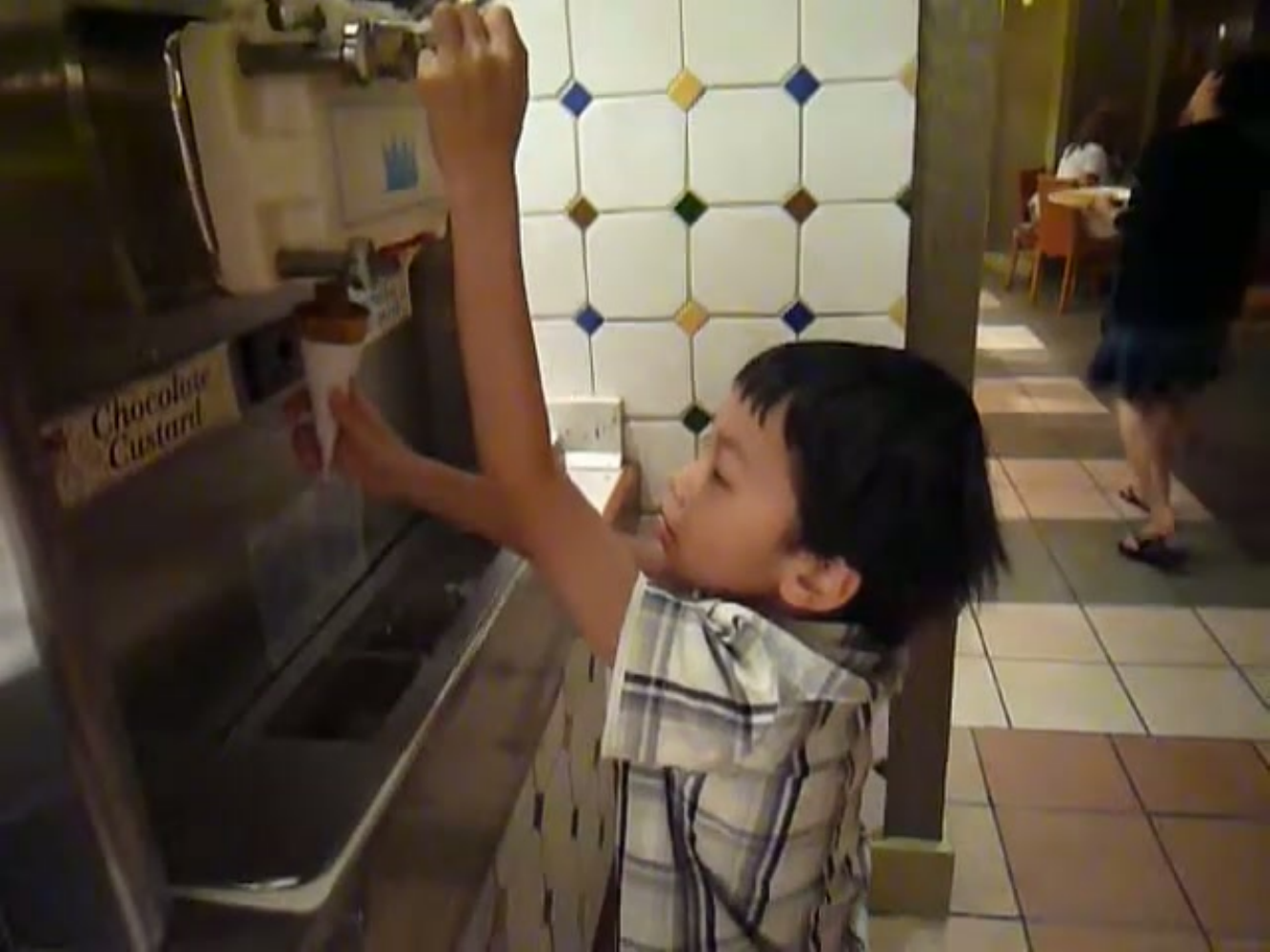}\hfill
\includegraphics[width=0.238\linewidth]{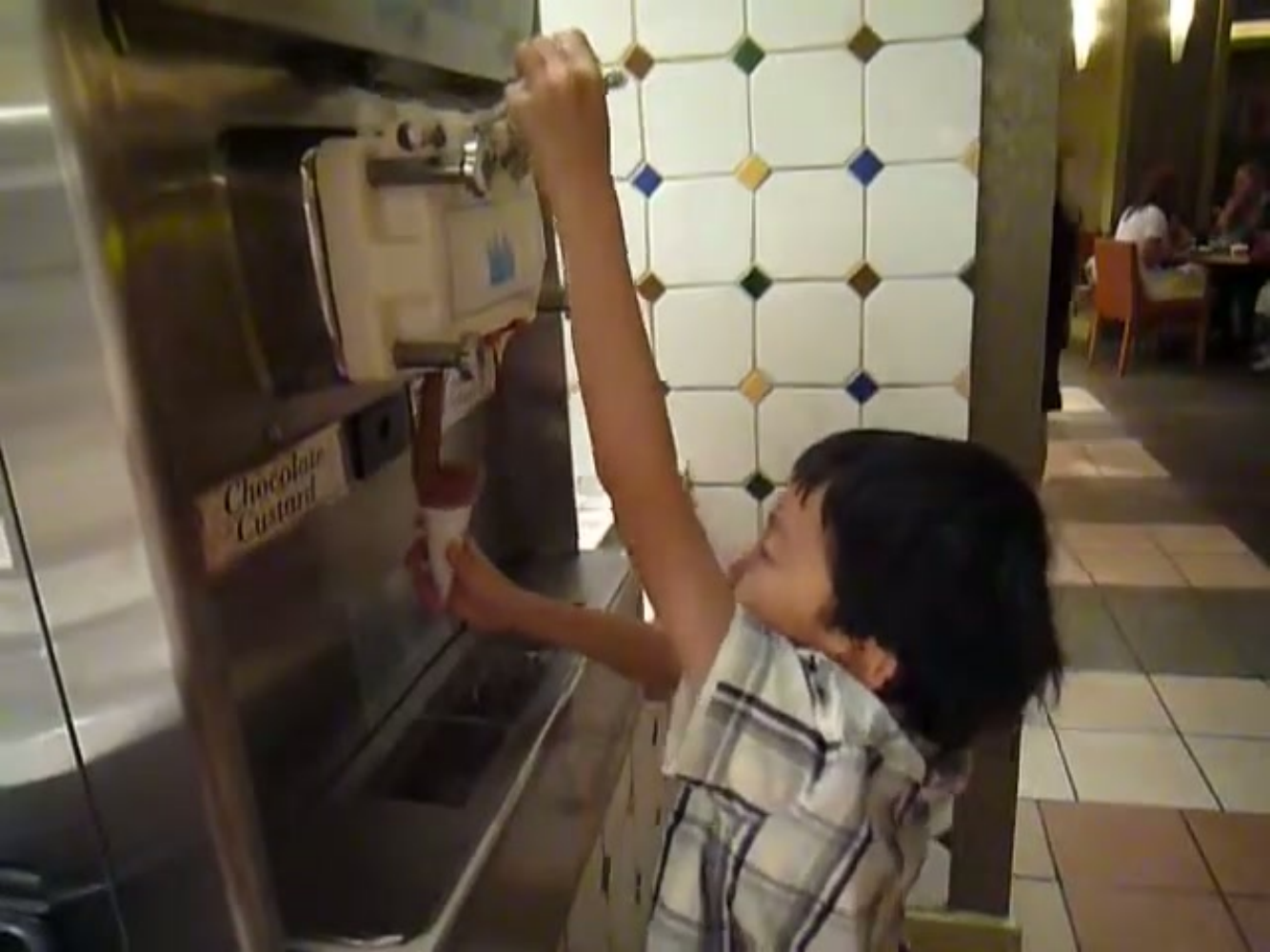}

\vspace{2mm}

\begin{tcolorbox}[
    colback=white,
    colframe=blue!35,
    boxrule=0.6pt,
    arc=1.5mm,
    left=2mm,
    right=2mm,
    top=1mm,
    bottom=1mm
]
\small
\RaggedRight
\textbf{Target reference.}
A big green garbage can is pushed up against a brick wall and a reindeer jumps out drinking from a coffee cup.
The reindeer continues to drink out of the coffee cup and eventually falls out of over the garbage can.

\vspace{1mm}
\textbf{Source reference.}
A little boy is filling an ice cream cone with ice cream.
He licks the ice cream cone.
He lifts the ice cream cone up to show the camera.

\vspace{1mm}
\textbf{Code-mined other-specific anchors.}
\textcolor{red!70!black}{cone; cream; ice; ice cream cone; enjoy cream; ice cream; capture with cream.}
\end{tcolorbox}

\vspace{1.5mm}

\begin{tcolorbox}[
    colback=white,
    colframe=blue!35,
    boxrule=0.6pt,
    arc=1.5mm,
    left=2mm,
    right=2mm,
    top=1mm,
    bottom=1mm
]
\small
\RaggedRight
\textbf{Original model summary for the claimed target slot.}
An animated deer climbs out of a green dumpster, holding an
\textcolor{red!70!black}{ice cream cone} with excitement.
The background features a brick wall and a street scene, adding context to the urban setting.
The deer attempts to \textcolor{red!70!black}{enjoy its ice cream}, even managing to catch some of it in its mouth playfully.
The scene is whimsical, capturing the deer's joyful interaction with the
\textcolor{red!70!black}{ice cream}.

\vspace{1mm}
\textbf{Human label.}
Confirmed cross-video leakage: the claimed target video shows a deer/reindeer drinking from a cup, while the highlighted ice-cream anchors are supported by another video in the same input group.
\end{tcolorbox}

\end{tcolorbox}
\caption{
\textbf{Manually verified prompt-placement leakage case.}
The automatic miner selected the highlighted concrete anchors because they are unsupported by the claimed target reference but supported by another video in the same input group. The human audit labels this case as confirmed cross-video leakage.
}
\label{fig:prompt_placement_leakage_deer_icecream}
\end{figure*}

\end{document}